\documentclass[acmtog]{acmart}

\acmSubmissionID{1024}

\AtBeginDocument{%
  }

\setcopyright{cc}
\setcctype{by}
\acmJournal{TOG}
\acmYear{2025} \acmVolume{44} \acmNumber{4} \acmArticle{} \acmMonth{8} \acmPrice{}\acmDOI{10.1145/3730913}

\usepackage{rotating}
\usepackage{soul}
\usepackage{colortbl}
\usepackage{tikz} 
\usetikzlibrary{decorations.fractals,spy}

\newcommand{\NEW}[1]{#1}
\newcommand{\myhref}[1]{\href{#1}{\color{blue}{#1}}}%

\renewcommand{\midrule}{\hline}
\renewcommand{\toprule}{\midrule\midrule}
\renewcommand{\bottomrule}{\midrule\midrule}

\definecolor{firstcolor}{RGB}{150,160,210} 
\definecolor{secondcolor}{RGB}{196,216,235} 
\newcommand{\first}[1]{\cellcolor{firstcolor}{#1}}
\newcommand{\second}[1]{\cellcolor{secondcolor}{#1}}
\newcommand{\rot}[1]{\rotatebox[origin=c]{90}{#1}}
\newcommand{\centeredtab}[1]{\setlength{\tabcolsep}{0pt}\begin{tabular}{c} #1 \end{tabular}}
\newcommand{\picwithtext}[3]{
	\centeredtab{\begin{tikzpicture}
			\node[anchor=south west, inner sep=0] at (0,0) {\includegraphics[width=#2]{#1}};
			\node[fill=black, fill opacity=0.5, anchor=south, inner sep=1pt, text opacity=1] at (0.5*#2,0.05) {\color{white}#3};\end{tikzpicture}}}

\newcommand{\dualspypic}[7]{
	\centeredtab{\begin{tikzpicture}[spy using outlines={rectangle,cyan,magnification=#7,size=0.485*#2, connect spies}]
			\node[anchor=south west, inner sep=0] at (0,0) {\includegraphics[width=#2]{#1}};
			\spy on (#3) in node [right] at (#4);
			\spy on (#5) in node [left] at (#6);
		\end{tikzpicture}}}

\begin{document}
\graphicspath{{figures/}}
\title{On-the-fly Reconstruction for Large-Scale Novel View Synthesis from Unposed Images}
\author{Andreas Meuleman}
\email{andreas.meuleman@gmail.com}
\author{Ishaan Shah}
\email{ishaan.n.shah@gmail.com}
\author{Alexandre Lanvin}
\email{laanvin@gmail.com}
\affiliation{
	\institution{Inria, Université Côte d'Azur}
	\country{France}
}
\author{Bernhard Kerbl}
\email{kerbl@cg.tuwien.ac.at}
\affiliation{
	\institution{TU Wien}
	\country{Austria}
}
\author{George Drettakis}
\email{George.Drettakis@inria.fr}
\affiliation{
	\institution{Inria, Université Côte d'Azur}
	\country{France}
}


\begin{abstract}
Radiance field methods such as 3D Gaussian Splatting (3DGS) allow easy reconstruction from photos, enabling free-viewpoint navigation. Nonetheless, pose estimation using Structure from Motion and 3DGS optimization can still each take between minutes and hours of computation after capture is complete. SLAM methods combined with 3DGS are fast but struggle with wide camera baselines and large scenes.  
We present an on-the-fly method to produce camera poses and a trained 3DGS \emph{immediately} after capture.
Our method can handle dense and wide-baseline captures of ordered photo sequences and large-scale scenes. To do this, we first introduce fast  initial pose estimation, exploiting learned features and a GPU-friendly mini bundle adjustment. 
We then introduce direct sampling of Gaussian primitive positions and shapes, incrementally spawning primitives where required, significantly accelerating training. These two efficient steps allow fast and robust joint optimization of poses and Gaussian primitives. 
Our incremental approach handles large-scale scenes by introducing scalable radiance field construction, progressively clustering 3DGS primitives, storing them in anchors, and offloading them from the GPU. Clustered primitives are progressively merged, keeping the required scale of 3DGS at any viewpoint. 
We evaluate our solution on a variety of datasets and show that it can provide on-the-fly processing of all the capture scenarios and scene sizes we target\NEW{. At the same time our method remains competitive -- in speed, image quality, or both -- with other methods that only handle specific capture styles or scene sizes. }
\end{abstract}



\begin{teaserfigure}
	\centering
    \setlength{\tabcolsep}{1pt}
    \renewcommand{\arraystretch}{0.6}
    \begin{tabular}{cccc}
        \centeredtab{\includegraphics[height=0.185\textwidth]{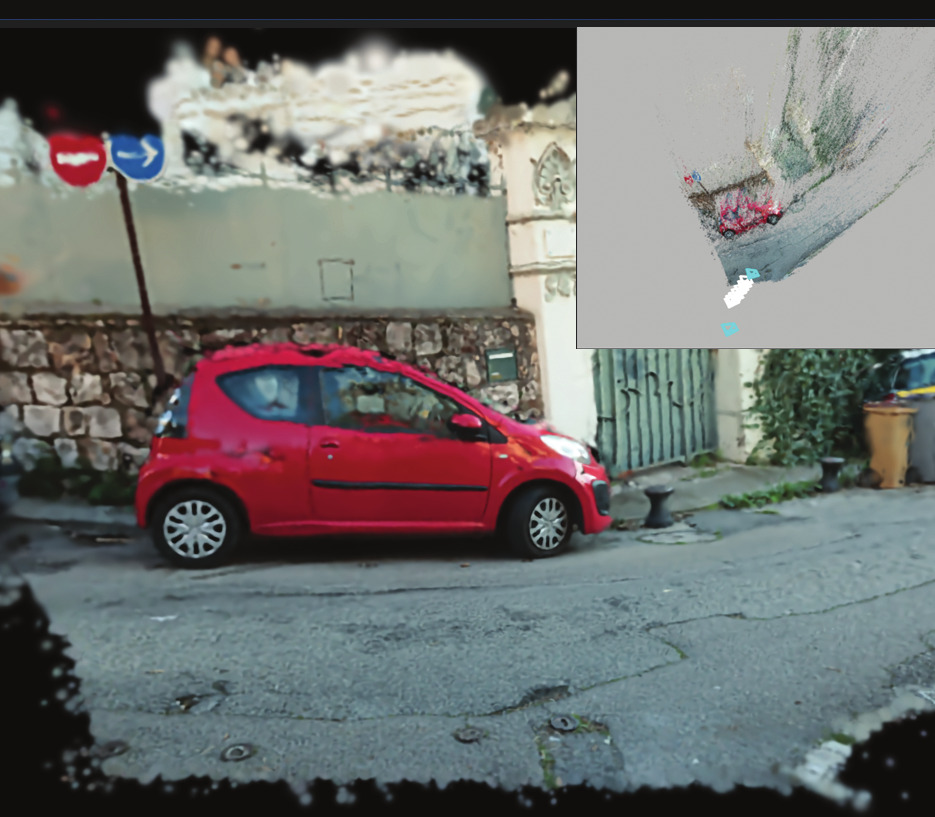}} & 
        \centeredtab{\includegraphics[height=0.185\textwidth]{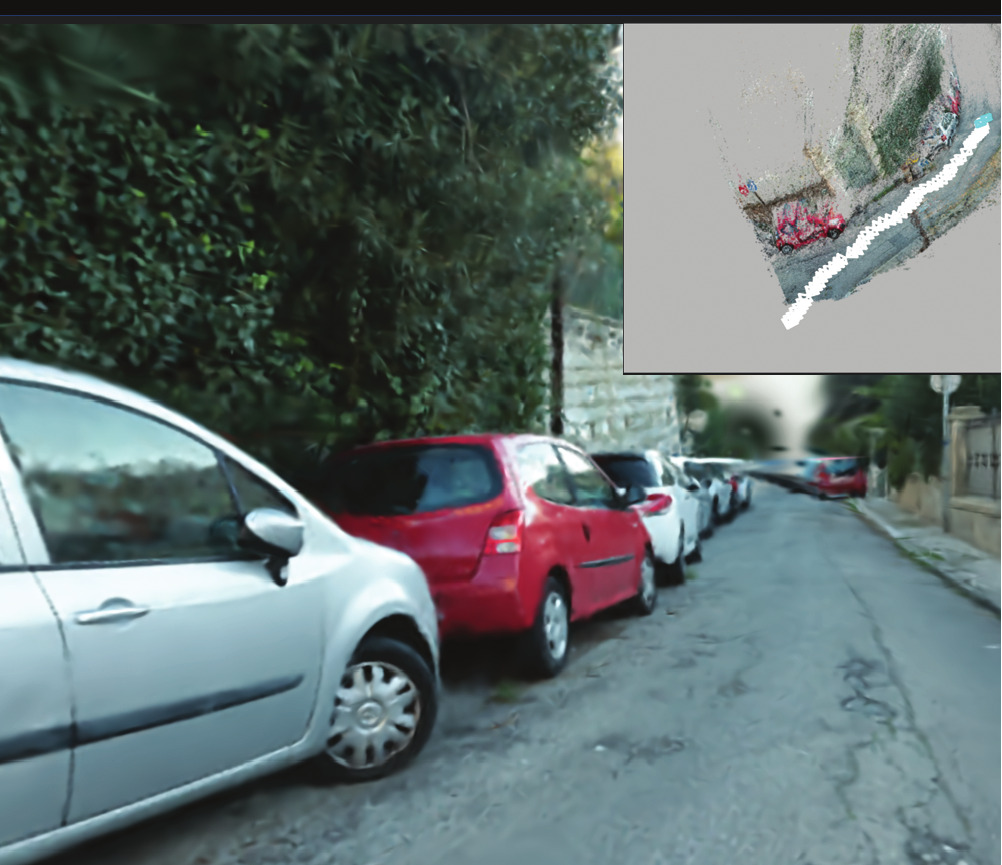}} & 
        \centeredtab{\includegraphics[height=0.185\textwidth]{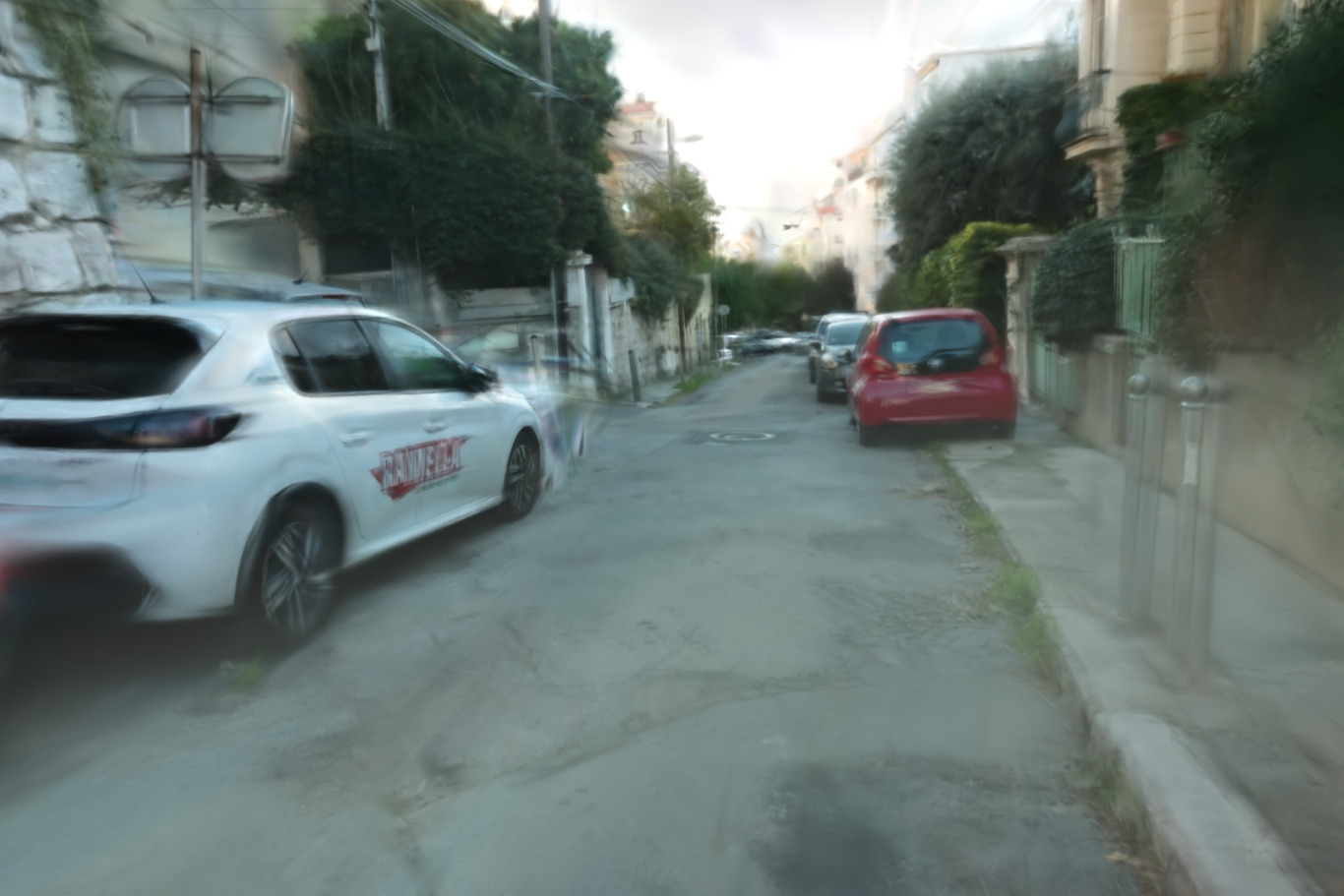}} & 
        \centeredtab{\includegraphics[height=0.185\textwidth]{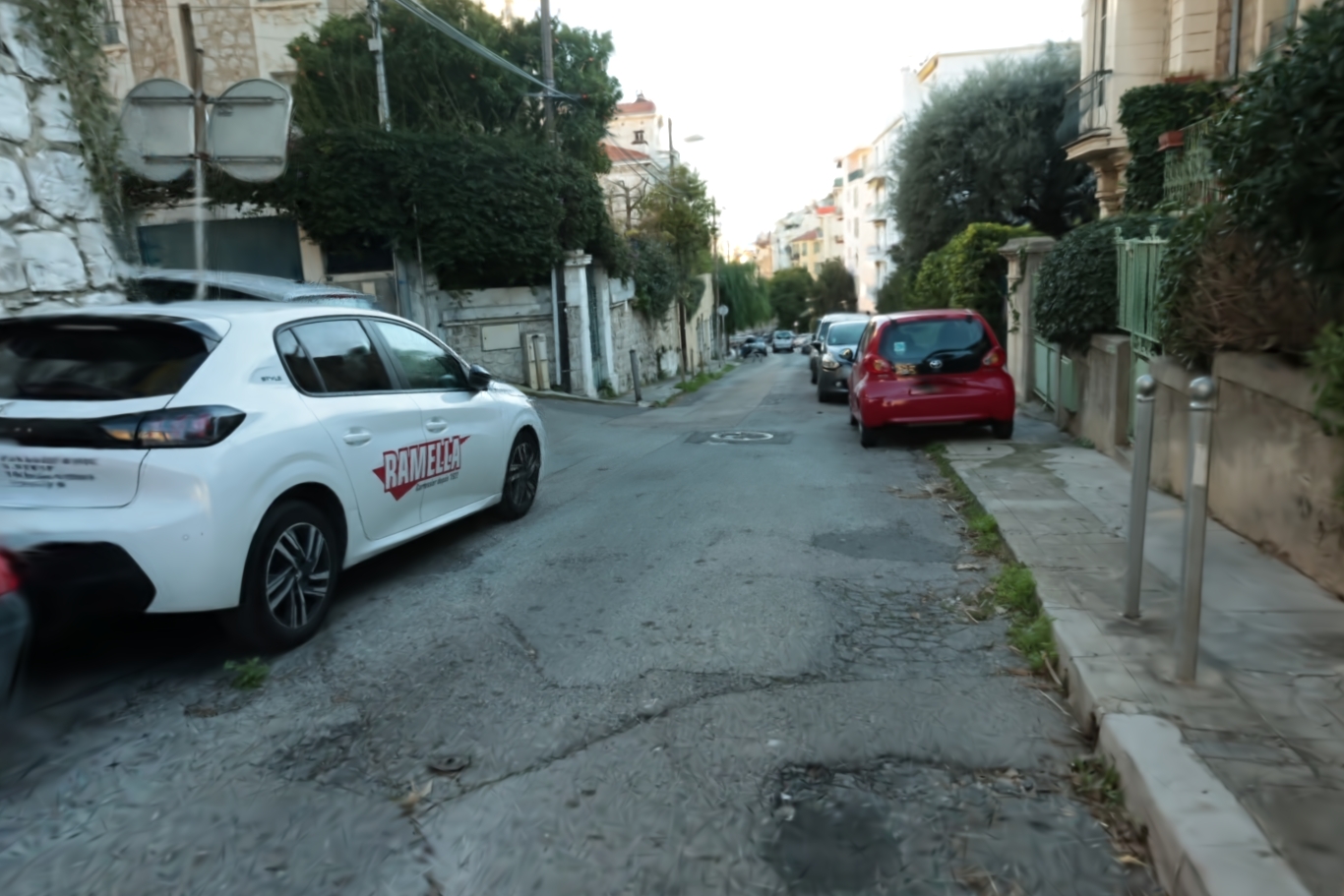}} \\
		(a) Initialization & (b) On-the-fly 3DGS & (c) H3DGS: 22hrs, PSNR 12~dB & (d) Ours: 25min, PSNR 21.7~dB
    \end{tabular}
	\vspace{-7pt}
	\caption{\label{fig:teaser}
Our method performs on-the-fly reconstruction from an unposed, ordered image sequence.
The total processing time for our method is 30min for this sequence of over 4000 images: with our method, the poses and radiance field are \emph{immediately available} after taking the photos. The scene was captured in a 1km walk in 30min, using a camera in "drive mode" taking photos at 3 images/sec, resulting in around 4000 images.
	Our method incrementally reconstructs a 3D Gaussian representation along with the camera poses. \NEW{In} the middle right, we show a novel view of the same scene reconstructed with Hierarchical 3DGS~\cite{hierarchicalgaussians24}, that requires 22 hours of processing for camera pose estimation and 3DGS optimization, in contrast to our method (right) that has completed all processing by the time photos are taken. The camera calibration of \NEW{hierarchical 3DGS} fails in many places later on the path, leading to novel view synthesis failure \NEW{in some places} (please see video and additional results at \myhref{https://repo-sam.inria.fr/nerphys/on-the-fly-nvs/}).
	} 
\end{teaserfigure}

\maketitle

\section{Introduction}

Radiance Field solutions for novel-view synthesis~\cite{barron2022mipnerf360,mueller2022instant,duckworth2023smerf} take a multi-view dataset of images as input and create a 3D digital version of a real scene, enabling free-viewpoint navigation.
Since the recent introduction of 3D Gaussian Splatting (3DGS)~\cite{kerbl3Dgaussians} --
a fast radiance field method with high visual quality -- there has been an explosion in the application of radiance fields in domains as diverse as e-commerce, extended reality, film and video, robotics, 3D generative modelling etc.  
Most such solutions first need to perform pose estimation for all cameras upfront~\cite{schoenberger2016sfm} followed by 3DGS optimization;
for typical wide-baseline datasets (tens of) minutes are still required for each of the two steps, even with the fastest methods~\cite{taming3dgs}.  
For larger scenes, these steps each require many hours of processing after capture has finished, hindering the adoption of such solutions in applications. We present a method for \emph{on-the-fly} radiance field construction that works on consumer hardware: with our solution, camera poses and the scene representation are immediately available by the time scene capture is finished.

While Structure-from-Motion (SfM)~\cite{schoenberger2016sfm} followed by 3DGS~\cite{kerbl3Dgaussians,taming3dgs} are too slow for on-the-fly processing, SLAM solutions are much faster~\cite{tosi2024nerfs}, but typically expect dense, video-like input, and often focus less on visual quality. These solutions often struggle with wide-baseline multi-view captures, typically used for radiance-field reconstruction.

Similarly, 3DGS optimization is still too costly for on-the-fly reconstruction. 
3DGS optimization starts from a sparse, noisy set of SfM points, while the \emph{densification} process provides additional points from heuristics. As a result, the optimization needs to proceed carefully, requiring many iterations. 
This results in long training times, even for the most efficient versions~\cite{taming3dgs}. Finally, while some solutions have been proposed to allow 3DGS for large scenes~\cite{octreeGS,hierarchicalgaussians24,liu2024citygaussian}, they all require the poses of all cameras to be estimated beforehand, for scene subdivision and/or hierarchy construction. 
Our goals are to provide both camera calibration and usable 3DGS reconstruction immediately after capture, either for dense or wide-baseline capture, and most importantly for large-scale scenes. Other methods have difficulty satisfying all of these goals. Note that we require captured images to be given in an ordered sequence, since we present an \emph{incremental} radiance field construction method. 

We observe that accurate pose estimation can be costly; if we relax the need for accuracy in a first step, we can reformulate pose estimation to be GPU-friendly and thus significantly faster even if the initial guess is approximate. 
For such an approach to be possible, joint pose/3DGS estimation needs to subsequently refine these poses. 3DGS differentiable rendering is well-suited to this task, since the gradients from the rendering loss can flow to the poses, improving their accuracy. Finally, for joint optimization to be effective, we also need to accelerate 3DGS; a good strategy to achieve this goal is to directly \NEW{sample} well placed Gaussians, reducing densification and making the optimization easier.

Given these observations, we first propose lightweight initial pose estimation that leverages learning-based matches with image neighbors. The matches and their currently optimized poses are used to quickly estimate the next camera pose. We also reformulate the problem to allow fast GPU-friendly mini bundle adjustment. Second, instead of densifying, we introduce a method for direct sampling of Gaussian primitives. When adding a new image, we directly choose the positions and sizes of 3D Gaussian primitives created, by estimating the probability that a given pixel should spawn a Gaussian, and also directly sample the size of each primitive. This greatly reduces the need for densification, providing the required optimization speed. Given our incremental pose estimation and training, 
joint optimization is accelerated and the risk of getting stuck in local minima is reduced. Finally, our efficient incremental optimization is naturally suited to a sliding-window clustering and merging approach which we propose, that stores parts of the scene as \emph{anchors} that are placed in space as image capture advances.
Taken together, these elements allow our method to incrementally process images in various capture styles, providing camera poses and 3DGS reconstruction on-the-fly. 

\noindent
In summary, our contributions are:

\begin{itemize}
	\item A fast initial pose estimation method based on deep feature matching and GPU-friendly mini bundle adjustment.
	\item Probability-based direct sampling of position and shape of Gaussian primitives, greatly alleviating the need for gradient-driven densification resulting in fast 3DGS optimization
Together with fast pose estimation, our sampling allows effective incremental joint optimization of poses and the radiance field.
	\item A sliding window clustering and merging strategy that allows on-the-fly processing of large-scale scenes using \emph{anchors}.
\end{itemize}

\noindent
We run evaluations on a variety of datasets, showing that our method is one of the only solutions that can provide on-the-fly processing of all the capture scenarios we target. 
\NEW{At the same time it remains competitive -- in speed, image quality, or both -- with other methods that only handle specific capture styles or scene sizes. }

\section{Related Work}

Novel View Synthesis (NVS) generates images of a scene from viewpoints not observed during capture, allowing free-viewpoint navigation~\cite{kerbl3Dgaussians,barron2022mipnerf360}. 
Our contributions are in pose estimation and radiance field optimization for NVS, in particular for large scenes. We briefly review the directly relevant literature for each of these domains, and refer the reader to more complete surveys, e.g., for 3DGS~\cite{chen2024survey3dgs,fei20243d} and SLAM with radiance fields~\cite{tosi2024nerfs}.

\paragraph{Camera pose estimation}
Pose estimation using SfM is most often a significant part of the overall computation to create the scene representation~\cite{kerbl3Dgaussians}, given its high computational expense \cite{schoenberger2016sfm, kopf2021robust, zhao2022particlesfm}. Despite recent advances~\cite{pan2024glomap, Wang_2024_CVPR, duisterhof2024mast3rsfmfullyintegratedsolutionunconstrained, brachmann2024acezero}, this process remains too slow for on-the-fly incremental reconstruction. 
SLAM approaches \cite{murTRO2015, Schops_2019_CVPR, ORBSLAM3_TRO} leverage frame-to-frame consistency to estimate camera poses incrementally but struggle with large baselines.
Learning-based SLAM methods \cite{DeepFactors, DeepV2D, homeyer2024droid}
improve robustness but are computationally intensive, which quickly becomes prohibitive when optimizing 3D Gaussians simultaneously with pose estimation. 
Recently, Spann3r \cite{wang20243d} achieved fast pose and point cloud prediction using a transformer-based approach, DUSt3R \cite{Wang_2024_CVPR}. However, these methods suffer from low-resolution output and substantial pose drift over longer videos.
In contrast, we propose a lightweight mini bundle adjustment as a first step by structuring the computation in a GPU-friendly manner; our initial pose estimation is thus fast and does not have to be highly accurate,
since we subsequently correct poses during our efficient joint optimization. 

\begin{figure*}[!h]
	\def\svgwidth{\linewidth}
	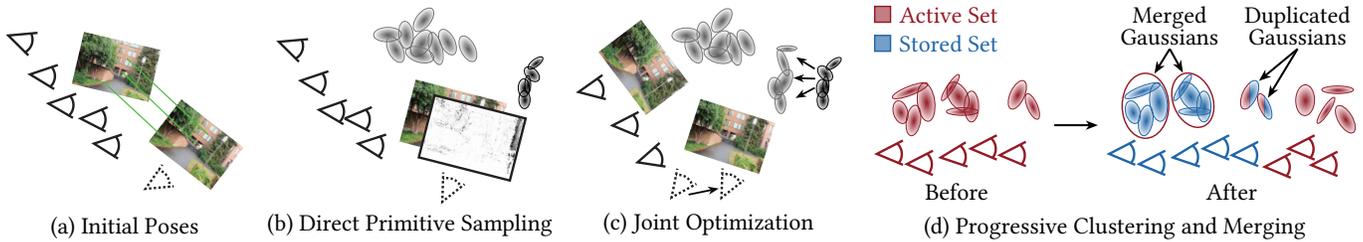\vspace{-7pt}
\caption{
\label{fig:overview} 
	Overview of our method. As each new image arrives, (a) we first use learning-based feature matching, reformulated in a GPU-friendly manner to achieve fast initial pose estimation (b) we perform direct sampling of Gaussian primitives by estimating the probability that a pixel should generate a Gaussian, and also sample the size, (c) the two first steps enable fast and effective joint optimization, 
improving both poses and scene representation and (d) we incrementally cluster into \emph{anchors}, and merge nodes allowing us to represent large-scale scenes.
\NEW{
When placing an anchor, the Gaussians from the active set are stored for later rendering. We create the new active set by merging primitives that appear small and duplicating the others, reducing the number of Gaussians being optimized.
}
}
\end{figure*}

\paragraph{Novel view synthesis}

Neural volumetric representations \cite{Lombardi:2019,mildenhall2020nerf,barron2021mipnerf} have enabled tremendous progress in the field, boasting high visual quality, but suffer from expensive optimization due to ray-marching and large neural networks. 
More explicit or hybrid representations \cite{yu_and_fridovichkeil2021plenoxels,mueller2022instant,Chen2022ECCV,barron2023zipnerf,hu2023Tri-MipRF,liu2024ripnerf} improve efficiency but remain relatively slow to render and optimize. 
Recently, 
3D Gaussian splatting \cite{kerbl3Dgaussians} represents the scene with 3D Gaussian primitives that can be efficiently projected and rasterized on the GPU. Each primitive stores opacity and Spherical Harmonics (SH) to represent appearance; these are blended together to recreate the appearance of the scene with high fidelity. 
3DGS achieves real-time rendering but requires careful initialization and densification, i.e., creating new primitives during optimization to achieve high-quality results. 
Recent \NEW{solutions} have improved densification and reduced primitive counts, thus speeding up the method with minimal quality loss \cite{taming3dgs, Reducing3Dgs, hhuang2024photoslam}. 
Several methods propose explicit initialization \cite{minisplatting}, but still require additional densification; unpublished concurrent work \cite{fang2024mini} refines this idea to further accelerate computation.
However, these methods are unsuited for on-the-fly incremental scene reconstruction
mainly because they require all camera poses to be computed beforehand, and densification requires many optimization iterations, increasing the overall cost. We propose direct sampling of Gaussian primitives alleviating the need for densification, reducing 3DGS optimization cost, and incremental joint optimization compatible with large scenes. 

\paragraph{Joint poses and 3D reconstruction}
Jointly optimizing camera poses and 3D scene representations is a natural approach to eliminate dependence on known camera poses \cite{scnerf2021, lin2021barf, liu2023robust, meuleman2023localrf, HF-SLAM, deng2024compact3dgaussiansplatting}. 
Such methods, however, require expensive optimization times—from seconds per frame to days per scene—and often rely on RGB-D data, e.g., \cite{HF-SLAM, deng2024compact3dgaussiansplatting}.
Near real-time performance has been achieved in recent work that jointly optimize poses and 3D Gaussians \cite{keetha2024splatam, peng2024rtgslam, yan2023gs, zhu2024_loopsplat}; Again, most of these require RGB-D input. A few methods handle RGB-only inputs \cite{gsslam2024, li2024GGRt}, but typically do not use matching-based pose initialization, increasing the number of required gradient-based optimization iterations, especially when the baseline is large. While some achieve real-time performance by reducing iteration counts \cite{yan2023gs, peng2024rtgslam}, they are suboptimal for datasets with large baselines (Sec.~\ref{sec:eval}).
Some methods also alternate between pose estimation and reconstruction, further increasing optimization time \cite{gsslam2024, Fu_2024_CVPR}. 

Some recent methods incorporate frame-to-frame matching to reduce optimization time, but can require hours for complete optimization~\cite{COGS2024}, or while improving speed with DUSt3R-based initialization, have quadratic matching time with the number of frames~\cite{fan2024instantsplat}.
Photo-SLAM combines ORB-SLAM3 with 3DGS, achieving real-time performance, but given ORB-SLAM3's limitations has difficulty with wide baselines, and requires additional computation time to sufficiently densify the representation.
In currently unpublished work, several methods propose SLAM solutions using 3DGS~(e.g., \cite{homeyer2024droid,hislam2,cartgs,zerogs}) with strategies such as monocular depth priors, or focusing more on loop closure.
None of these methods supports large-scale scenes, while maintaining both interactive pose estimation and acceptable visual quality.  

In contrast to these solutions, our fast initial pose estimation and Gaussian sampling steps strikes a better balance of computational load, simultaneously handling SLAM-like and wide-baseline capture, as well as large-scale scenes.

\paragraph{Representations for large-scale scenes}
Some NVS methods handle general captures \cite{zhang2020nerfanalyzingimprovingneural,yu_and_fridovichkeil2021plenoxels,barron2022mipnerf360} or naturally extend to larger scenes \cite{kerbl3Dgaussians}, but have insufficient capacity for very large trajectories. 
Several large-scale representations subdivide scenes into local radiance fields \cite{Turki_2022_meganerf,Block-NeRF,meuleman2023localrf, duckworth2023smerf, xu2023gridguided, mi2023switchnerf}, often requiring specific handling of seams~\cite{duckworth2023smerf,Block-NeRF}.
Similar solutions have been developed for 3DGS, requiring divide-and-conquer solutions~\cite{lin2024vastgaussian} followed by involved steps to construct and adapt a hierarchy to the data~\cite{octreeGS,hierarchicalgaussians24,liu2024citygaussian}. 
All of these training approaches are incompatible with incremental reconstruction, since they require prior knowledge of the full scene camera poses. 
Progressive optimization of poses and local radiance fields has been explored \cite{meuleman2023localrf}, but optimization is slow, requiring up to 40 hours for 1000 images.
Pose estimation methods struggle with large datasets, as shown by COLMAP's~\cite{schoenberger2016sfm} failure to produce consistent trajectories in certain public large-scale datasets~\cite{meuleman2023localrf,hierarchicalgaussians24}. Hierarchical 3D Gaussians \cite{hierarchicalgaussians24} also need time-intensive, per-chunk bundle adjustment, taking up to hours per chunk.

Our efficient joint optimization overcomes all these limitations and is inherently suited to incremental scene construction, using a sliding window solution for large-scale scenes.

\section{Overview}

We propose an \emph{on-the-fly} method to estimate camera poses and compute a complete radiance field at the speed of photo acquisition, designed for large scenes. 
Our method has four main components: 1) A fast -- but approximate -- initial pose estimation, adopting a careful design to allow GPU-friendly mini bundle adjustment; 2)
A direct sampling method to find the position and shape of Gaussian primitives, by estimating the probability of each pixel to generate a Gaussian, significantly reducing the need for densification; 3) 
A method for joint optimization of poses and 3DGS that is very efficient, thanks to the first two steps, improving the initial versions of both poses and the radiance field; 4) An online scalable optimization using a sliding set of anchors that progressively clusters 3DGS primitives in space, allowing the treatment of large-scale scenes.
Figure~\ref{fig:overview} provides an overview of these steps.

\section{Method}
\label{sec:method}

\subsection{Lightweight Initial Pose Estimation}
\label{sec:poseinit}

We first compute approximate initial poses, that will later be improved by joint
optimization;
our design choices thus prefer speed over flexibility.
Specifically, to fully exploit the GPU, we first use a limited number of keypoints, reducing expensive memory access on the GPU and second, we formulate our solution as a fixed-size problem thus exploiting GPU core parallelization. Initial pose estimation thus has three stages: feature extraction, bootstrapping and subsequent frame estimation. 

\paragraph{Feature extraction.}
A fast feature keypoint detector and descriptor~\cite{potje2024xfeat} is applied to each input image, generating 6144 keypoints per frame. 

\paragraph{Bootstrapping}
We first wait until the first $N_\mathrm{init}$ frames have arrived,
then run exhaustive matching between each pair of these ($N_\mathrm{init}$ is 8 in our experiments). 
From this set of matches, we optimize the focal length, poses and 3D keypoints' positions by minimizing the reprojection error. 
Following standard practice, we implement this mini bundle adjustment as Levenberg-Marquardt optimization~\cite{LevenbergLMOpt,MarquardtLMOpt,LMopt}. 
Our mini bundle adjustment is lightweight and efficient, compared, e.g., to full 3DGS rendering SGD optimization used by other methods~\cite{gsslam2024}. 

The key to our efficient solver is to carefully layout the problem so that each 3D point is seen from a fixed number of images. 
This results in a fixed-size sparse Jacobian $J$ of the reconstruction error, which is easy to build and enables an efficient solve method on the GPU. Specifically,
we compute the Jacobian of the reprojection error with respect to the camera pose $J_{\text{cam}}$ and 3D point position $J_{xyz}$. At each iteration, similar to standard solvers, we compute the reprojection error and its Jacobian $J$, that is built from $J_{\text{cam}}$ and $J_{xyz}$, which are both sparse.  
Since we fix the non-zero block sizes, we can pre-allocate memory and compute every block independently with fixed-size computation, allowing us to exploit batch-processing on the GPU. This simplified layout avoids the need for flexible solvers such as Ceres \cite{ceres} that are typically used for bundle adjustment.

\paragraph{Pose Estimation for Subsequent Frames}
In each new frame, we match its keypoints 
to those in the last $N$ registered frames ($N$ is 6 in our experiments). To establish 3D-2D correspondences, we estimate the 3D positions of keypoints in each of the $N$ past frames using the known (previous) camera poses and triangulation. If this estimation fails, we use rendered depth.
We then estimate the camera pose and inliers from these 3D-2D correspondences using GPU-parrallel RANSAC with our mini bundle adjustment as estimator. 
After this initialization, we run 20 iterations of the mini bundle adjustment with all inliers to refine the pose.
Finally, a 3D Gaussian primitive is created for each triangulated keypoint.
\NEW{Although a keypoint can be seen from many images due to transitive matches, we restrict supervision to the last $N$ registered frames to maintain a fixed-size problem.}

To ensure the method recovers in challenging scenarios (pure rotations, scale drift), we rerun the bootstrapping when the mean distance between the last twenty cameras is below 0.1/3. 
If the projection error is below 1 pixel, we update the $N_\text{init}$ last frames' poses by aligning it to the previously estimated ones.

\subsection{Sampling Gaussian Primitives}
\label{sec:gaussinit}

\begin{figure*}[!h]
    \centering 
    \setlength{\tabcolsep}{1pt}
    \renewcommand{\arraystretch}{0.6}
    \begin{tabular}{cccccc}
        \centeredtab{\rotatebox[origin=c]{90}{First init.}} & 
        \centeredtab{\includegraphics[width=0.19\textwidth]{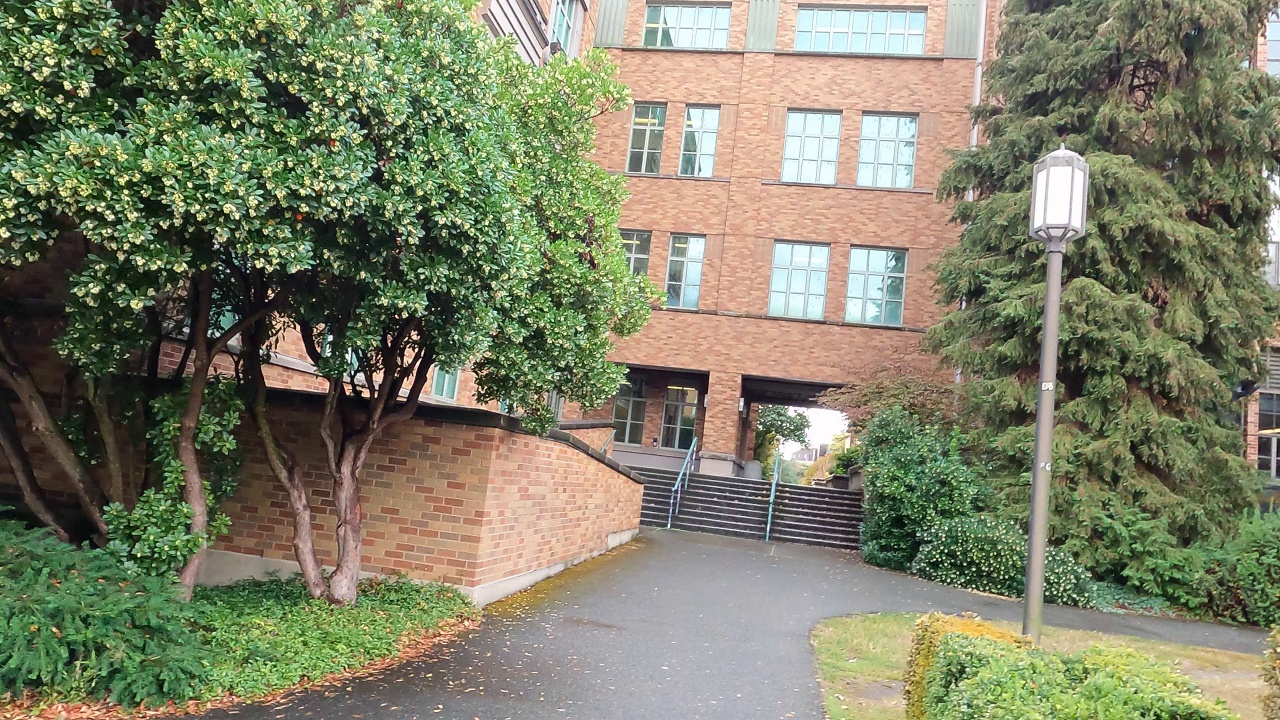}} & 
        \centeredtab{\includegraphics[width=0.19\textwidth]{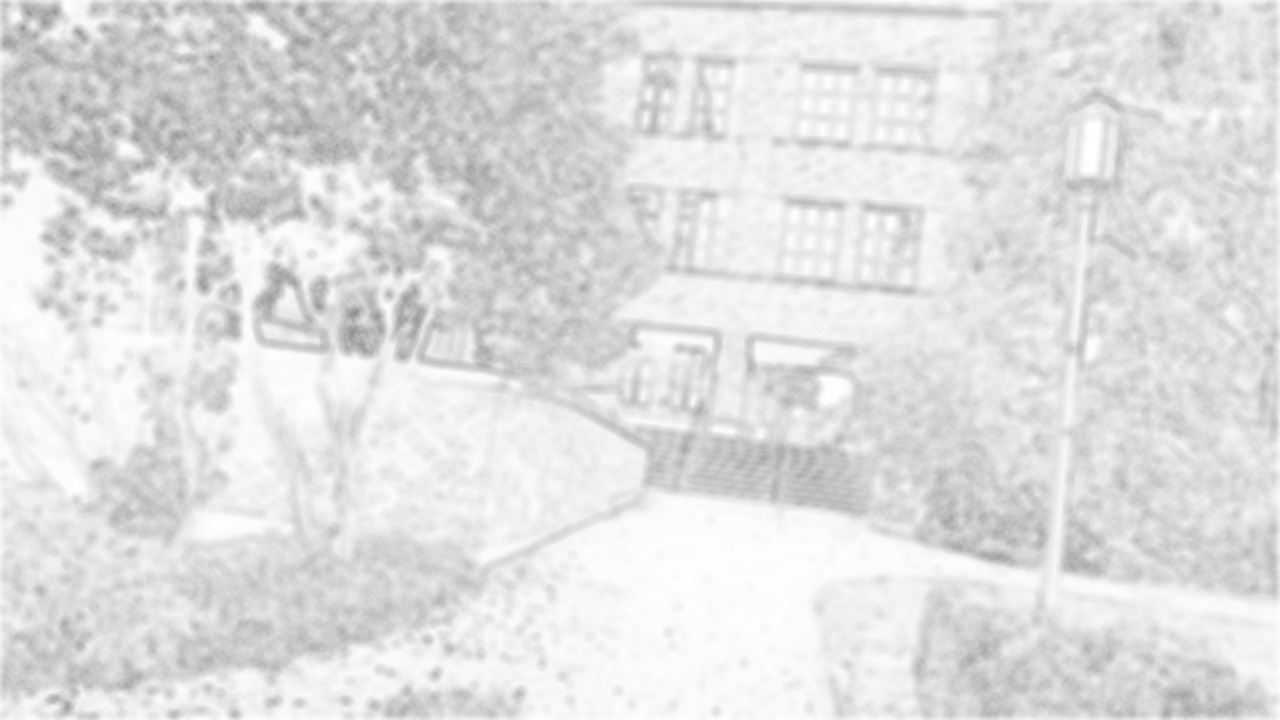}} & 
        \centeredtab{\includegraphics[width=0.19\textwidth]{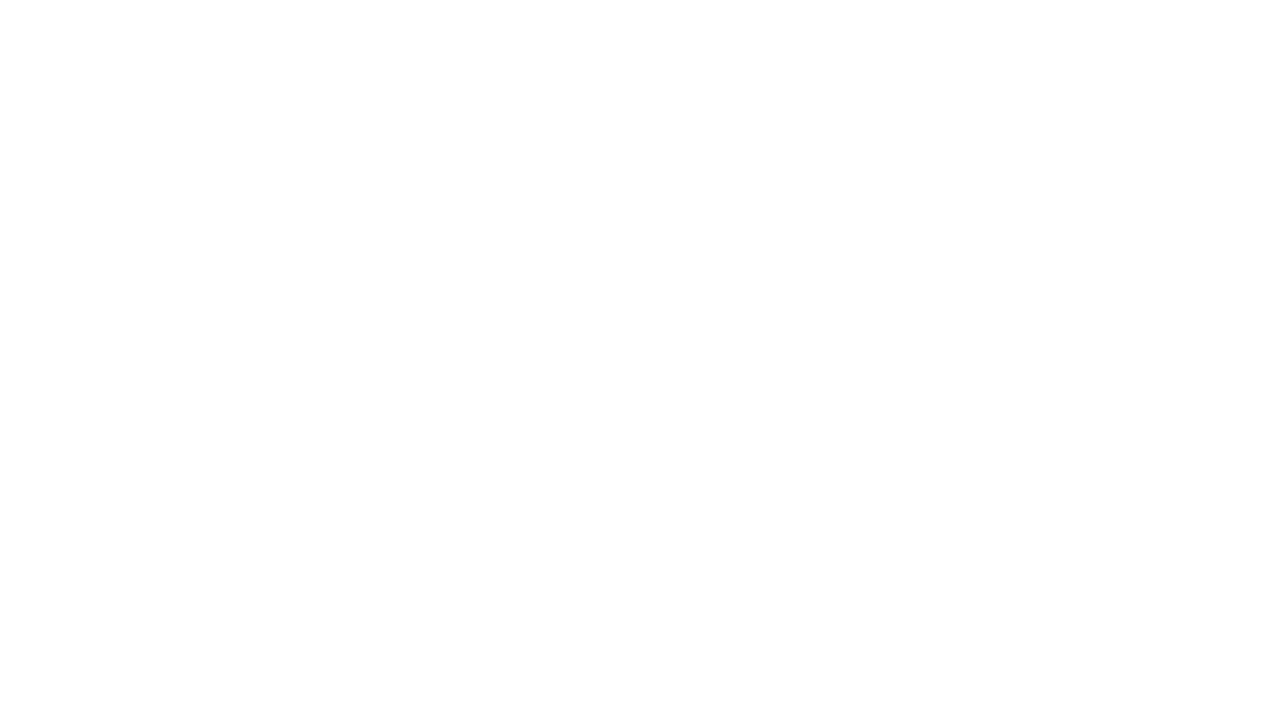}} & 
        \centeredtab{\includegraphics[width=0.19\textwidth]{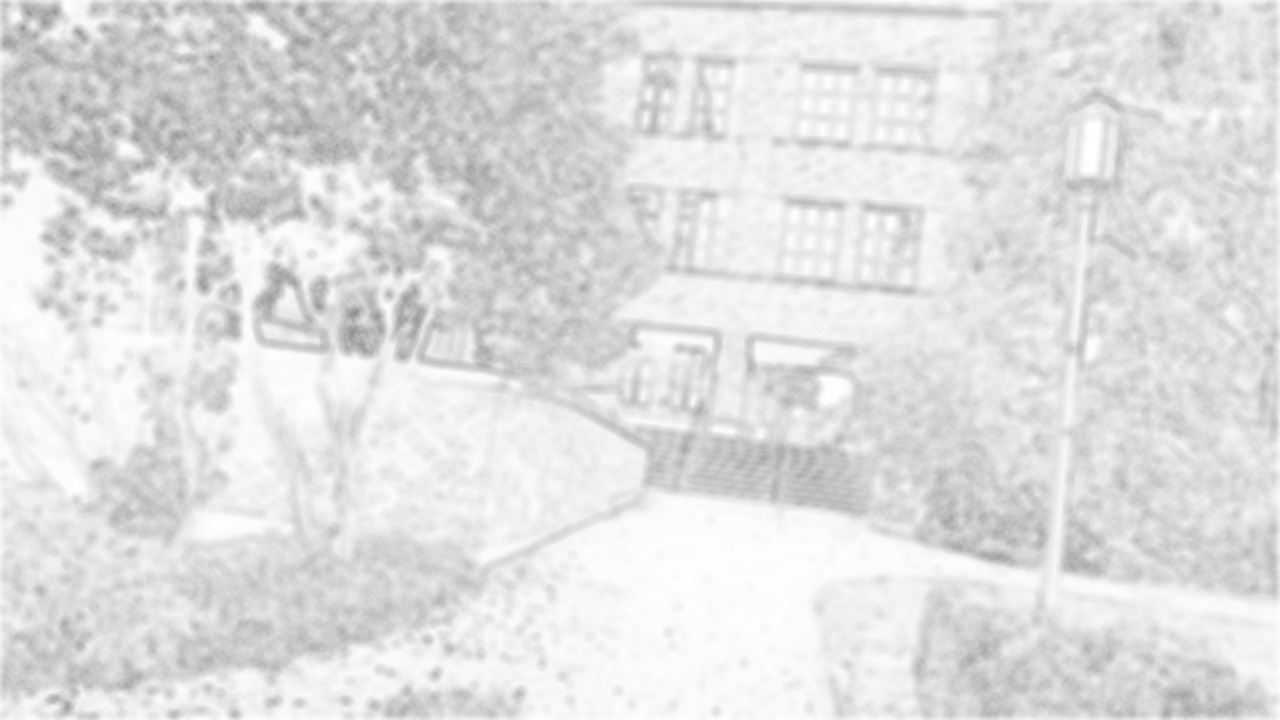}} & 
        \centeredtab{\includegraphics[width=0.19\textwidth]{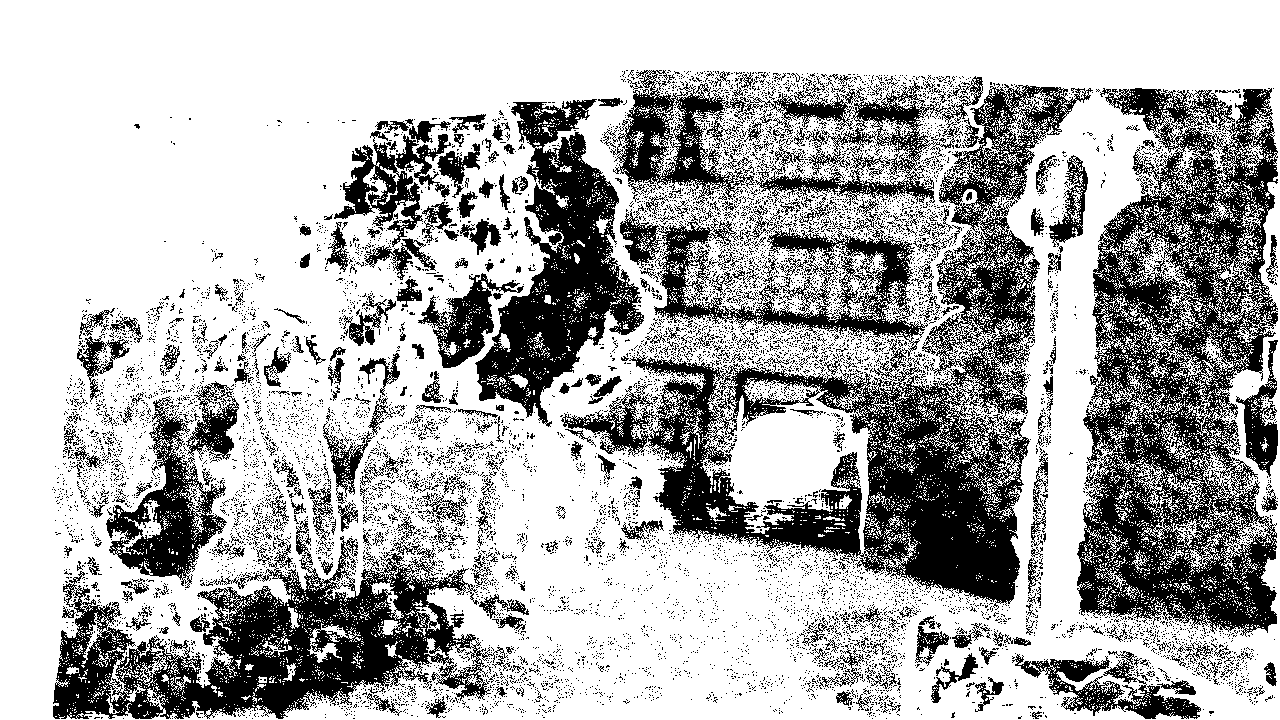}} \\
        \centeredtab{\rotatebox[origin=c]{90}{Later init.}} & 
        \centeredtab{\includegraphics[width=0.19\textwidth]{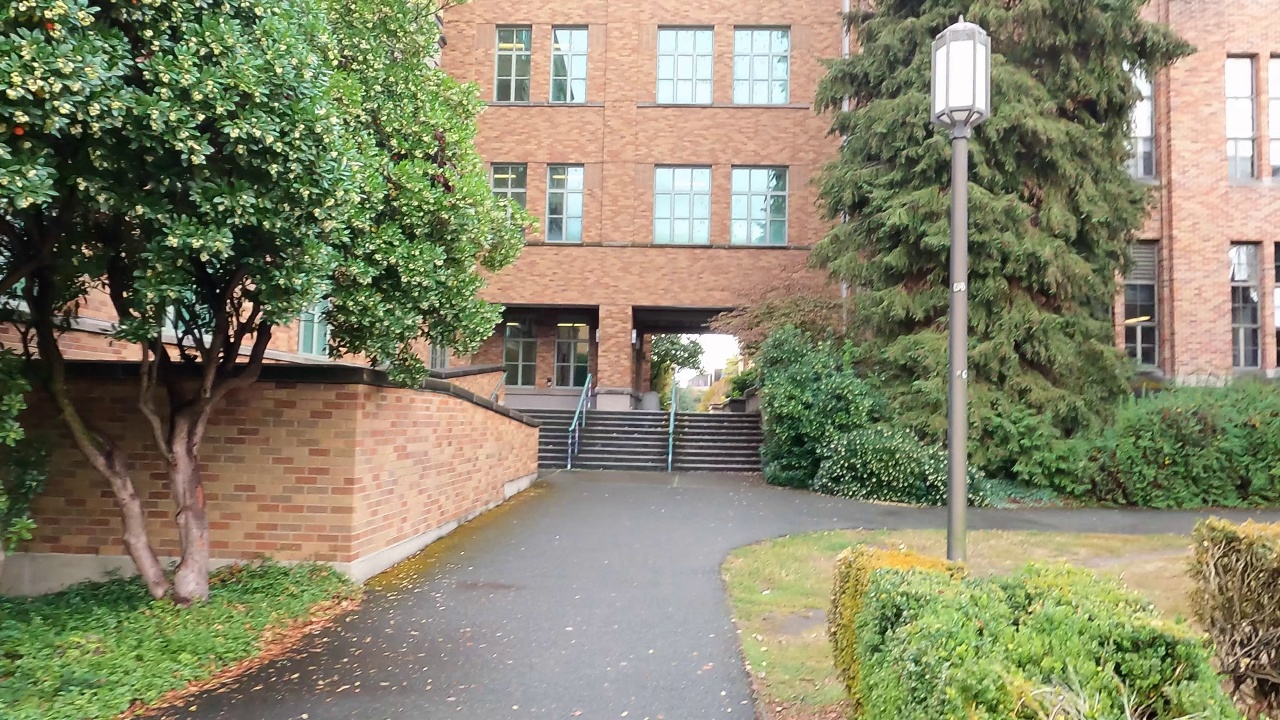}} & 
        \centeredtab{\includegraphics[width=0.19\textwidth]{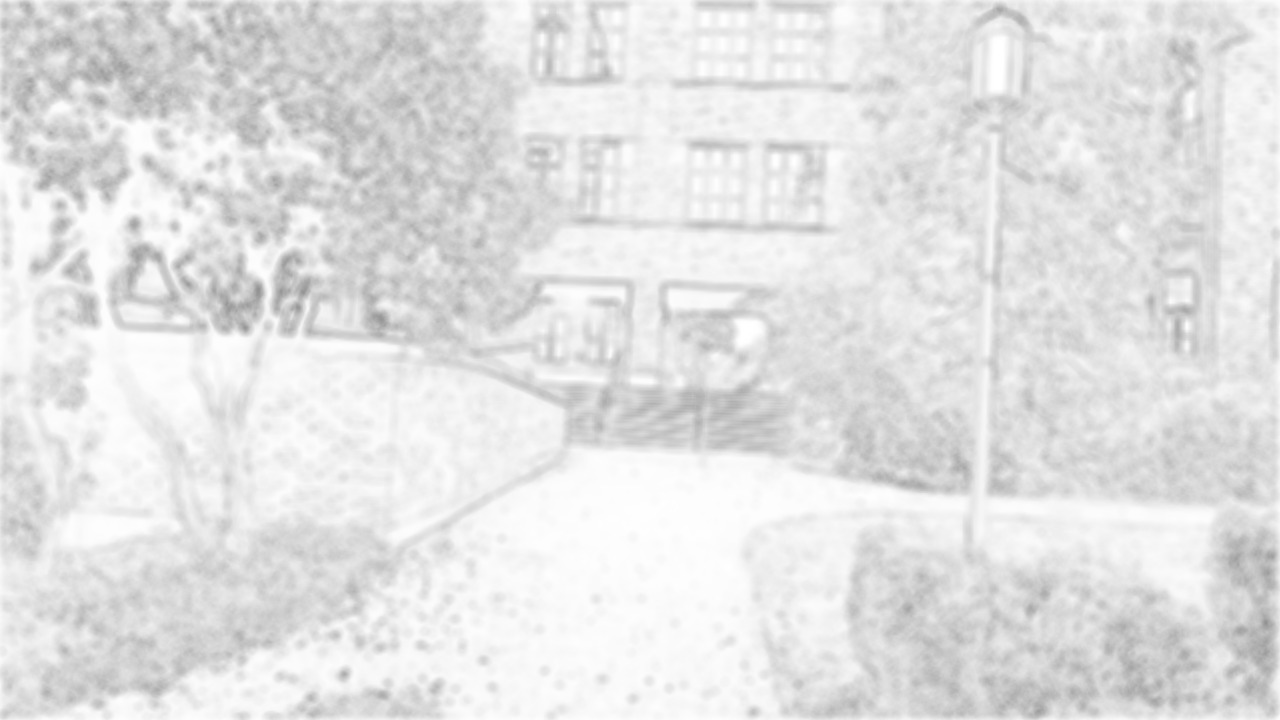}} & 
        \centeredtab{\includegraphics[width=0.19\textwidth]{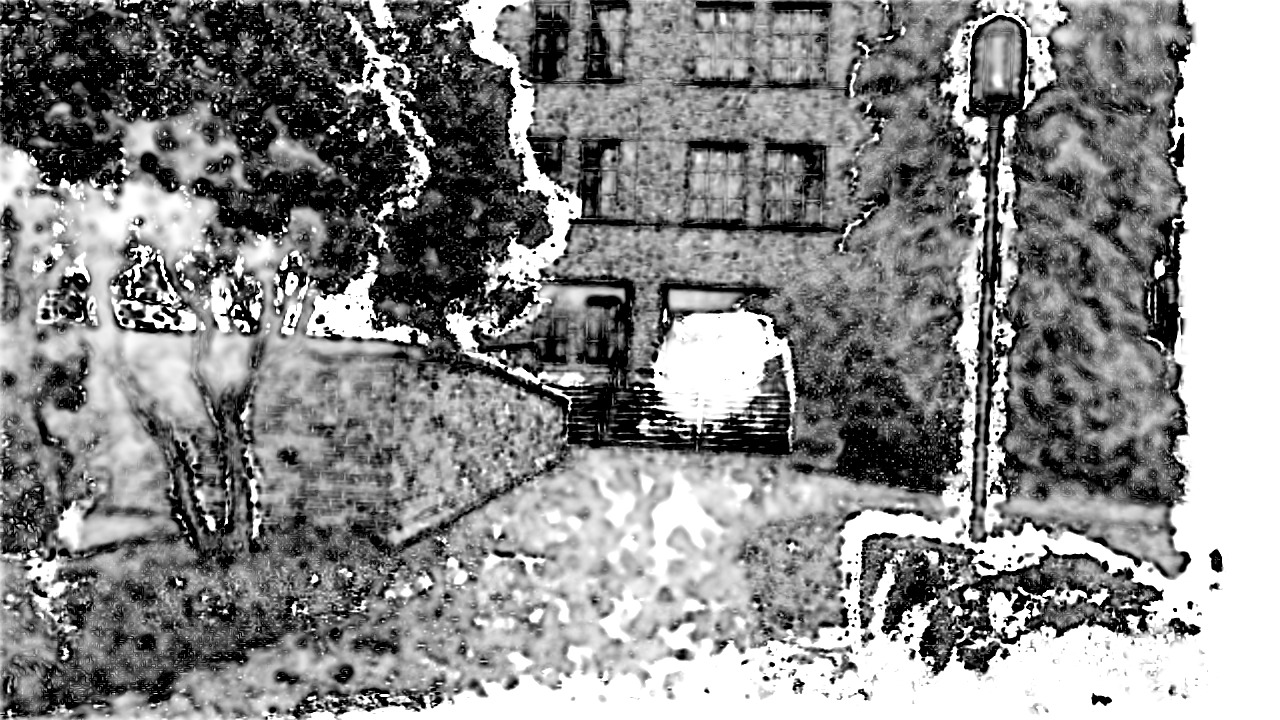}} & 
        \centeredtab{\includegraphics[width=0.19\textwidth]{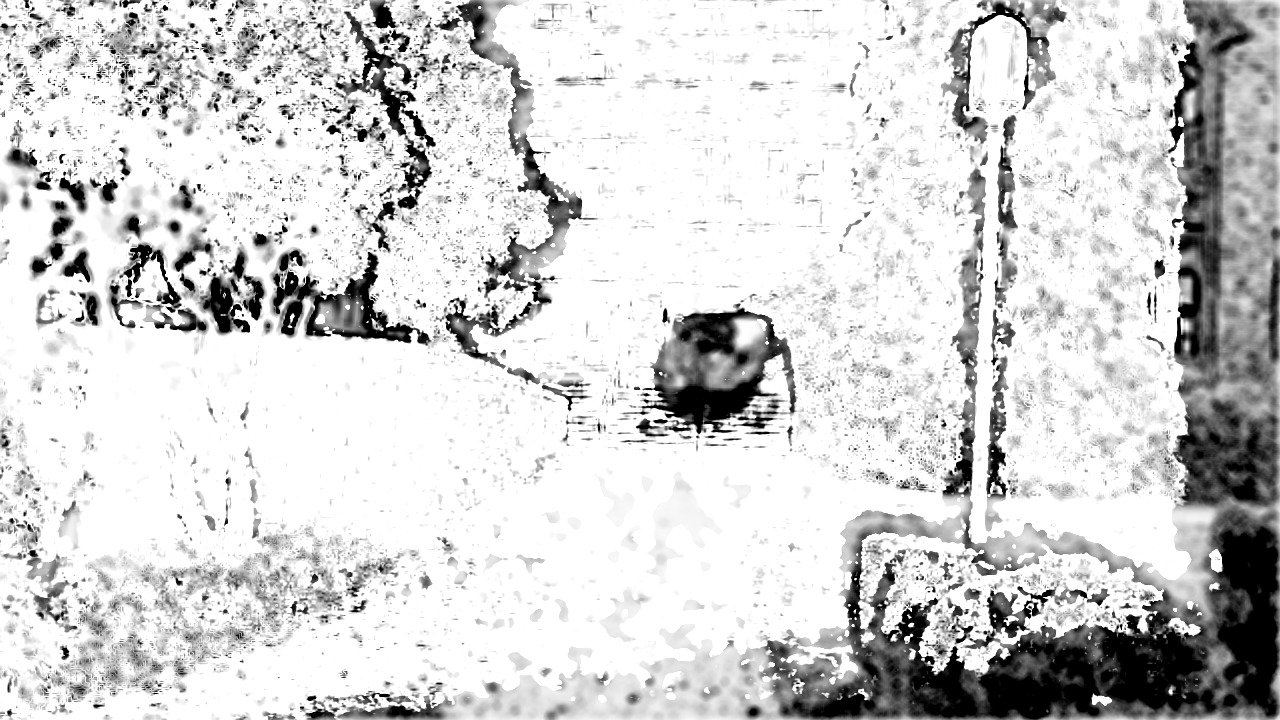}} & 
        \centeredtab{\includegraphics[width=0.19\textwidth]{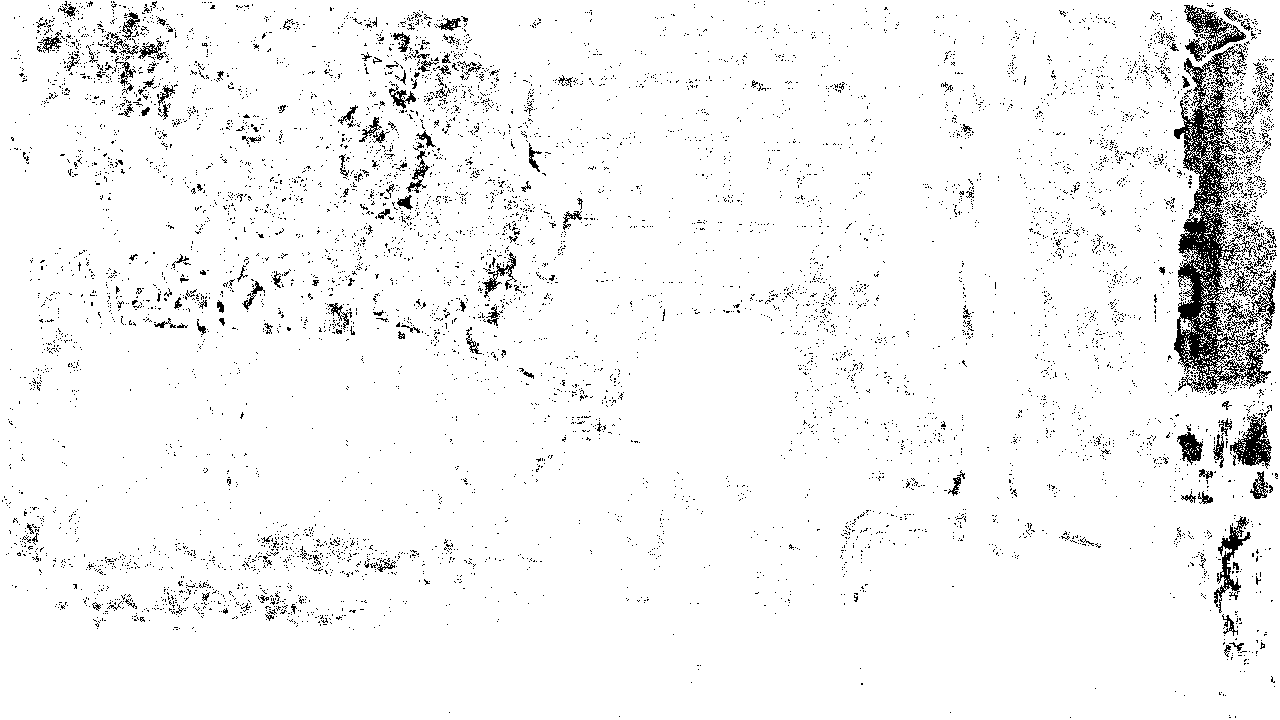}} \\
	    & (a) Input & (b) $P_L$ & (c) $\tilde{P}$ & (d) $P_s$ & (e) Sample mask
    \end{tabular}
    \caption{
        \label{fig:sampling}
	Direct sampling to place new primitives during joint optimization: When adding a new frame (a), we compute an intitial probability map $P_L$ based on the Laplacian of Gaussian (LoG) norm (Eq.~\eqref{eq:prob1}) (b). We then compute a penalty map $\tilde{P}$ based on the LoG norm of the rendered image (c); for the first image, this is empty, since there are no Gaussians to render. The final probability map $P_s$ is computed by subtracting the penalty from the initial probability (Eq.~\eqref{eq:prob_sampling}, (d)). From this probability, we create the mask to place new Gaussians (e).
        We observe that this method places more Gaussians in high-frequency areas (leaves), while texture-less regions feature less samples (road). Also, while the first initialization places primitives over the full image, the penalty map helps to avoid placing Gaussians in already well-represented areas. For example, only the unseen right part 
	of the later initialization has many new Gaussians. 
    }
\end{figure*}

To avoid the overhead and shortcomings of densification, we introduce 
a direct sampling method for Gaussian primitives.
At each frame, we sample a dense set of 3D Gaussian primitives satisfying two requirements: 1) place primitives to cover previously unseen regions, or to add additional detail in coarsely reconstructed parts of the scene and 2) avoid placing more primitives than actually required in any given region.
To satisfy these requirements simultaneously, we introduce a sampling method based on the probability that a pixel of a new frame should generate a primitive. \NEW{The steps of our sampling method are illustrated in Fig.~\ref{fig:sampling}.}

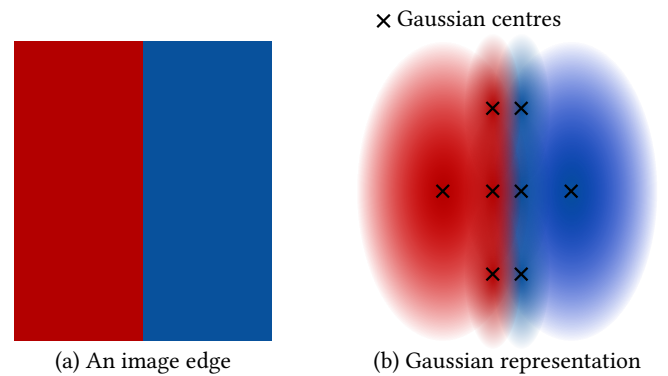
\begin{figure}[!h]
	\def\svgwidth{\columnwidth}
\begingroup%
  \makeatletter%
  \providecommand\color[2][]{%
    \errmessage{(Inkscape) Color is used for the text in Inkscape, but the package 'color.sty' is not loaded}%
    \renewcommand\color[2][]{}%
  }%
  \providecommand\transparent[1]{%
    \errmessage{(Inkscape) Transparency is used (non-zero) for the text in Inkscape, but the package 'transparent.sty' is not loaded}%
    \renewcommand\transparent[1]{}%
  }%
  \providecommand\rotatebox[2]{#2}%
  \newcommand*\fsize{\dimexpr\f@size pt\relax}%
  \newcommand*\lineheight[1]{\fontsize{\fsize}{#1\fsize}\selectfont}%
  \ifx\svgwidth\undefined%
    \setlength{\unitlength}{243.34868183bp}%
    \ifx\svgscale\undefined%
      \relax%
    \else%
      \setlength{\unitlength}{\unitlength * \real{\svgscale}}%
    \fi%
  \else%
    \setlength{\unitlength}{\svgwidth}%
  \fi%
  \global\let\svgwidth\undefined%
  \global\let\svgscale\undefined%
  \makeatother%
  \begin{picture}(1,0.56167618)%
    \lineheight{1}%
    \setlength\tabcolsep{0pt}%
    \put(0.19991438,0.00690365){\color[rgb]{0,0,0}\makebox(0,0)[t]{\lineheight{1.25}\smash{\begin{tabular}[t]{c}(a) An image edge\end{tabular}}}}%
    \put(0.76632843,0.00690365){\color[rgb]{0,0,0}\makebox(0,0)[t]{\lineheight{1.25}\smash{\begin{tabular}[t]{c}(b) Gaussian representation\end{tabular}}}}%
    \put(0,0){\includegraphics[width=\unitlength,page=1]{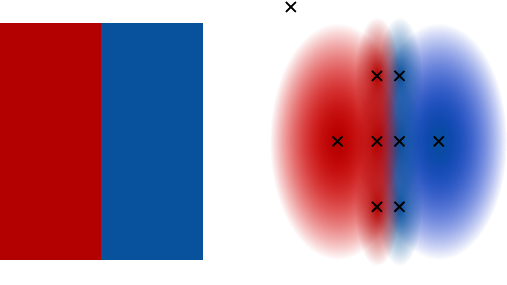}}%
    \put(0.59500799,0.54047617){\color[rgb]{0,0,0}\makebox(0,0)[lt]{\lineheight{1.25}\smash{\begin{tabular}[t]{l}Gaussian centres\end{tabular}}}}%
  \end{picture}%
\endgroup%

    \vspace{-15pt}
	\caption{\label{fig:gs_around_edge}
	To represent an image edge (a) accurately, more Gaussians should be placed on both sides of the discontinuity (b), placing them where they are required.
    }
\end{figure}

Existing Gaussian splatting SLAM approaches typically handle Gaussian initialization by either uniformly distributing Gaussians across the image \cite{yan2023gs} or placing them at keypoints \cite{hhuang2024photoslam}. 
However, uniform placement fails to adapt to the specific features of the input image, while keypoints alone tend to be too sparse, requiring further densification with the resulting increase in optimization time and number of primitives. 

We define the probability that a primitive should be generated at a given pixel based on two criteria:
1) 3D Gaussian primitives should be concentrated in areas with high-frequency details which cause discontinuities, and 2) 
Gaussians should be positioned on both sides of each discontinuity, to accurately represent edges (see Fig.~\ref{fig:gs_around_edge}).

\begin{figure}[!ht]
	\centering
	\def\svgwidth{0.7\columnwidth}
	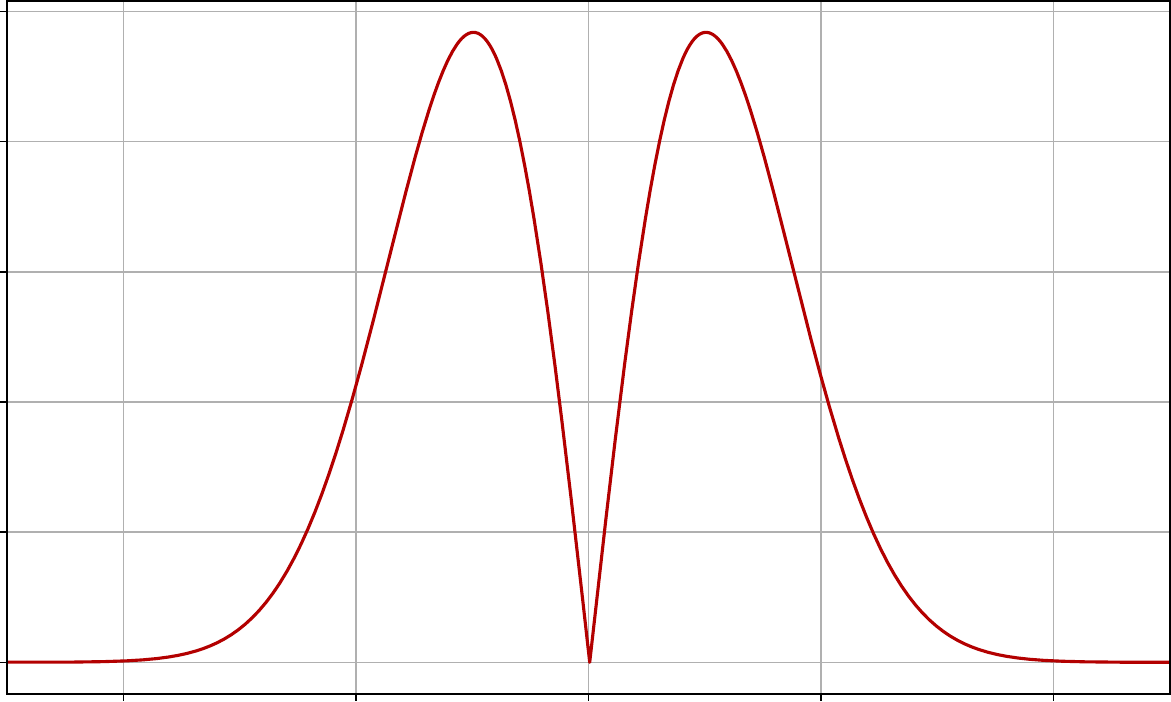
    \vspace{-7pt}
	\caption{\label{fig:logresp}
    The response of the norm of the Laplacian of Gaussian (LoG) function to a step function.
	The LoG operator highlights edges by producing two peaks on either side of a discontinuity. 
    }
\end{figure}
The probability that a given pixel should generate a primitive is thus based on the local spatial gradient. Using the norm of the Laplacian of Gaussian (LoG) operator~\cite{haralock1991computer} as a proxy for our probability fulfills both criteria: areas with high-frequency details yield high LoG norm, and a sharp edge produces two peaks on either side of the discontinuity (see Fig.~\ref{fig:logresp}). 
Thus, we assign an initial probability $P_L$ that a primitive should be generated at pixel $(x, y)$ based on the LoG norm:
\begin{equation}
    \label{eq:prob1}
    P_L(x, y) = \min\left(\left\|\nabla^2(n_\sigma) * I (x, y)\right\|, 1\right).
\end{equation}
Here, $I$ is the input image, and $n_\sigma$ is a Gaussian kernel with a standard deviation of $\sigma$.

To satisfy our second requirement, i.e., avoid placing excess Gaussians in areas \NEW{where there are already sufficient primitives to represent the edges,}
we render a view $\tilde{I}$ from the viewpoint of the new frame. 
We then compute the same quantity as in Eq.~\ref{eq:prob1} but for the rendered image $\tilde{I}$, providing a pixel-wise penalty $\tilde{P}$ that reduces the probability of placing new Gaussians in already reconstructed areas: 
\begin{equation}
    \label{eq:pen}
    \tilde{P}(x, y) = \min\left(\left\|\nabla^2(n_\sigma) * \tilde{I} (x, y)\right\|, 1\right).
\end{equation}
\noindent
The quantity $\tilde{P}$ will be similar to $P_L$ in regions where content has already been reconstructed, thus the final probability for adding a Gaussian at pixel $(x, y)$ is given by:
\begin{equation}
    \label{eq:prob_sampling}
    P_s(x, y) = \max\left(P_L(x, y) - \tilde{P}(x, y), 0\right).
\end{equation}
\noindent
This will reduce the probability to spawn primitives in regions that are already well represented by the rendered image and thus the representation.

\paragraph{Depth for Gaussian Primitive Positions}
\label{par:initpos}
With a set of pixel positions selected for conversion into 3D Gaussians, 
we now have a set of pixels that will spawn 3D Gaussian primitives;
the next step is to estimate their depths. 
We use Depth-Anything-2~\cite{depth_anything_v2} to estimate monocular depth, which we \NEW{align to the triangulated matches} using the same procedure as described in Kerbl et al.~\shortcite{hierarchicalgaussians24}. We then estimate depth using a standard correlation volume approach centered around the monocular depth. \NEW{This guided matching is essential as the monocular depth can exhibit significant errors.} Details of this computation are given in the Appendix.

\begin{figure}[!h]
    \centering 
    \setlength{\tabcolsep}{1pt}
    \renewcommand{\arraystretch}{0.6}
    \begin{tabular}{ccc}
        \centeredtab{\rotatebox[origin=c]{90}{Input View}} & 
        \centeredtab{\includegraphics[width=0.47\columnwidth]{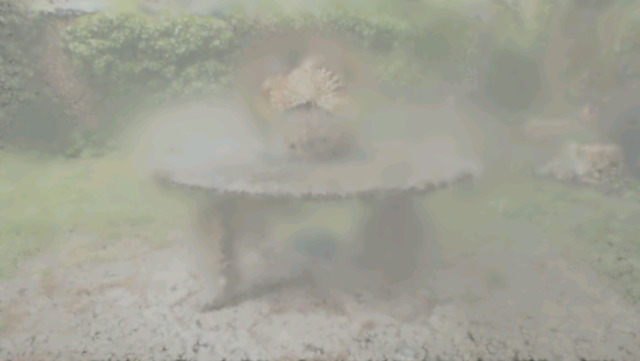}} & 
        \centeredtab{\includegraphics[width=0.47\columnwidth]{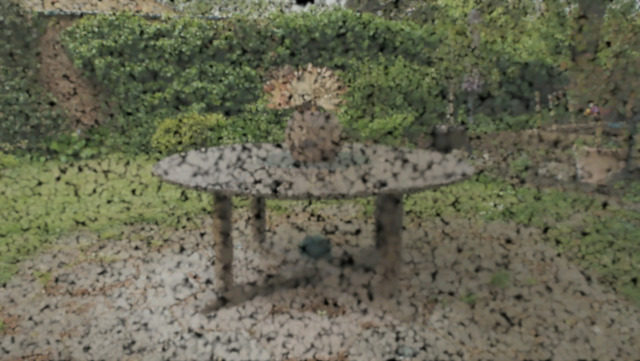}} \\
        \centeredtab{\rotatebox[origin=c]{90}{Novel View}} & 
        \centeredtab{\includegraphics[width=0.47\columnwidth]{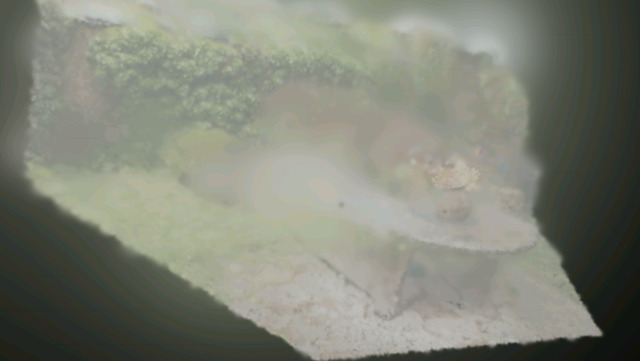}} & 
        \centeredtab{\includegraphics[width=0.47\columnwidth]{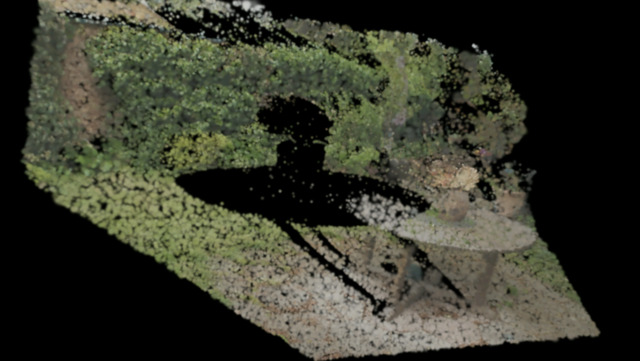}} \\
        & (a) Init. with 3D distance & (b) Init. w/ expected distance
    \end{tabular}
    \caption{
        \label{fig:initscale}
        Scale initialization. (a) With the initial scale based on the 3D distance to the 3-nearest neighbors, some Gaussians are too large. (b) Scale initialization based on the expected distance to the nearest neighbor is more appropriate, fitting the input image better. 
    }
\end{figure}

\paragraph{Primitive Size Parameter}
In 3DGS, the scale of the primitives is initialized based on the average distance to the approximate 3D 3-nearest neighbors. However, this approach tends to produce overly large Gaussians around discontinuities and is sensitive to outliers, resulting in an initialization that poorly matches the input frame (see Fig.~\ref{fig:initscale}).

To address this, we first estimate an appropriate scale in image space. Using the probability from Eq.~\eqref{eq:prob1}, we compute the expected distance to the nearest neighbor assuming a local 2D Poisson process of intensity $P_L(x, y)$ around the pixel $(x, y)$~\cite{clarkevans1954nn}:
\begin{equation}
    s' = \frac{1}{2\sqrt{P_L(x, y)}}.
\end{equation}
This calculation leverages the probability $ P_L $ before the penalty term, as a high penalty would imply that many Gaussians are already present.

Next, we convert from pixel space to 3D space using the camera’s focal length $ f $ and the estimated depth $ z $ for the pixel:
    $s = \frac{z s'}{f}.$
This approach provides an appropriate scale without requiring a nearest neighbor search, making it efficient. 
We then assign $ s $ to each dimension of the 3D Gaussian's scale vector $ S $.

\subsection{Joint Pose and Gaussian Optimization and Scheduling}
\label{sec:posegauss}

For each new image, we now have an initial estimation of poses and directly sampled positions and sizes of Gaussian primitives. These two fast initial steps enable efficient joint optimization of the poses and the 3DGS representation. 

For each new image received, we only register the frame if 
the median displacement of the keypoints exceeds $3\%$ of the screen width; these
registered frames are the \emph{keyframes}.
This ensures that only frames with meaningful parallax are used, improving geometry estimation and avoiding redundant frames. 

For each registered image, we run 30 Gaussian splatting optimization iterations, using the fast backpropagation and sparse-Adam optimizer from \cite{taming3dgs} to enhance iteration speed.

Learning rates are assigned per Gaussian, with decay rates adjusted based on the point at which each Gaussian was introduced.
Camera poses are optimized jointly using a 6D rotation representation for rotations \cite{6dRotation}. These poses receive gradient updates from Gaussian position and rotation optimization, but not from spherical harmonics, as propagating gradients through view-dependent color information can degrade pose quality \cite{liu2023robust}.

To capture low-frequency scene details first, 
accelerate optimization and avoid local minima, we employ a coarse-to-fine strategy \cite{yan2023gs,hhuang2024photoslam}. Specifically, every time an image is added, it is used for training with $2^l$ downsampling ($l~=~3$ in our experiments). Then, every five iterations, we decrement $l$ until we reach full image size. \NEW{We use proper filtering~\cite{Yu2024MipSplatting} to ensure correct multi-scale training. }While we do not perform densification, we do apply opacity culling as in the original 3DGS, removing primitives with very low opacity.

Our initial pose estimation prevents local minima when optimizing poses and Gaussians jointly, allowing us to handle wider-baseline than SLAM-based methods.
The incremental nature of our approach is well-suited to handling large environments, using our scalable incremental method we describe next.

\subsection{Scalable Incremental Gaussian Construction}
\label{sec:scalable}

Given our incremental pose and radiance field optimization, our final goal is to allow processing of large-scale environments. Previous solutions for large environments~\cite{octreeGS,hierarchicalgaussians24,liu2024citygaussian} incur significant overhead of creating, optimizing and maintaining hierarchical data structures, often requiring many hours for large scenes. This overhead is unacceptable for our goal of on-the-fly radiance field construction.

As images are processed, we maintain a set of \emph{Active Gaussian} containing primitives currently being optimized and rendered.
After some time, primitives placed and optimized earlier may appear very small or even sub-pixel-sized from the current camera location, contributing little to the rendered images. These primitives 
are offloaded from GPU to CPU RAM and stored at an \emph{anchor}. 
This gives a scene representation as a set of clusters that can be loaded back onto the GPU when required.
The clustering process has three steps: 1) detecting when to create an anchor, 2) clustering and primitive merging 3) Incremental optimization with a sliding window.

\paragraph{Detecting when to create an anchor.}
We define the size $\mathcal{S}$ of a primitive from a camera $i$ as $S / D$ where $D$ is the distance of the primitive center to the camera, and $S$ is the scale of the Gaussian.
		When we are at camera $i$ in the sequence, we check if more than 40\% of the \emph{Active Gaussians} have a size $\mathcal{S} < \tau_{\mathrm{min}}$ from camera $i-1$'s point of view ($\tau_{\mathrm{min}} = 1$ pixel). If they do, we trigger an update to create an anchor and merge these Gaussians.

\paragraph{Clustering and Primitive Merging.}
The set of \emph{Active Gaussians} before the update is copied to the newly created anchor.
The anchor stores a location, the set of Gaussian primitives, their optimization state and the keyframes they were optimized with. 

We next merge Gaussians that have a minor contribution to obtain a coarser representation of distant regions. 
To do this, we randomly select $\frac{1}{k+1}^\text{th}$ of primitives deemed too fine in the detection step. 
We then find the $k$ nearest neighbors for each of the selected primitives \NEW{using the method of Papantonakis et al.}~\shortcite{Reducing3Dgs} and merge them following the approach adopted by Kerbl et al~\shortcite{hierarchicalgaussians24} ($k~=~3$).
All other primitives are kept unchanged.

\paragraph{Sliding window incremental optimization.}
The merging process leaves us with a coarser representation of the scene, which becomes the new \emph{Active Gaussian} set. This set is optimized in the next iteration. 
Subsequent clustering steps will create new anchors, and distant content will become progressively coarser due to merging.
At the end of the capture path, the \emph{Active Gaussian} set is stored in a last anchor.
The use of anchors is illustrated in the supplemental video.

Once we have created the full scene representation of the scene, where different scale representations are stored with anchors, we can navigate freely in space. To perform rendering of novel views with our representation, we select the closest anchor to the camera's current position and render the Gaussians it contains.
When two anchors have a similar distance to the camera, we blend the Gaussians from both. 
If the two closest anchors from the camera's point of view are at distances $d_1$ and $d_2$ (with $d_1 \le d_2$), we define the overlap parameter as $o \in (0,0.5)$ (e.g., $o=0.1$) and compute a ratio $r = \frac{d_1}{d_2}.$
If $r \leq 1 - o$, then the blending weight for the closest anchor is $1$ and $0$ for the other anchor. Otherwise, we linearly blend the weight:
\begin{equation}
w(r) =
\begin{cases}
1, & \text{if } r < 1 - o, \\[1em]
1 \;-\; \bigl(r - (1 - o)\bigr)\,\frac{0.5}{o}, & \text{otherwise}.
\end{cases}
\end{equation}

\section{Results and Evaluation}

We implemented our method building on the 3DGS codebase~\cite{kerbl3Dgaussians}, adding a python based interactive viewer for training and online visualization after optimization. Source code of our method and the viewer, as well as additional material are available at \myhref{https://repo-sam.inria.fr/nerphys/on-the-fly-nvs/}.

Our method is robust to different capture styles, ranging from SLAM-like dense video to the more wide-baseline captures typically used for NVS. To demonstrate this, we evaluate on datasets taken with a variety of different capture styles.
For SLAM-like capture, we evaluate on the densely captured \textsc{TUM} dataset~\cite{TUMdataset}, often used for evaluation of SLAM-based approaches.
For intermediate, somewhat wider baseline and larger scale capture we evaluate on \textsc{Static Hikes} \cite{meuleman2023localrf}. 
For NVS wide-baseline capture, we test on a selection of scenes from the \textsc{MipNeRF360} dataset~\cite{barron2022mipnerf360}. 
For large-scale scenes, 
\NEW{we evaluate on the \textsc{SmallCity*} and \textsc{Wayve*} scenes, adapted from H3DGS~\cite{hierarchicalgaussians24} by using only the front camera.}
We have selected scenes from these datasets that have ordered image sequences, which is a requirement for our method. Finally, we have also captured a large dataset \textsc{CityWalk} with a Canon EOS R6 camera in drive mode at 3 images per second. 
The average number of images for the \textsc{TUM} datasets (fr1, fr2, fr3) is \NEW{2289, for \textsc{MipNeRF360} (garden, counter, bonsai) 239  and for \textsc{StaticHikes} (forest1, forest2, university2) 972. For the \textsc{H3DGS} the average is 2285}; our self-captured \textsc{CityWalk} scene has 4055 images, but has by far the largest spatial length of 1.1km.
We use image resolution of 1200-1600 width in all tests, unless stated otherwise, that usually corresponds to half resolution of the raw input data.

We ran all tests and evaluations on a workstation with an Intel Core i9 14900K CPU, 128GB of RAM, and an NVIDIA RTX 4090 GPU, or if we use a different configuration (e.g. when the method requires more GPU memory) we scale the timings to this setup by running 1000 calls to our CUDA rasterizer on each machine. \NEW{We use the same set of parameters for all scenes.}

\begin{table*}
\setlength{\tabcolsep}{3pt}
\caption{\label{tab:slam-eval} \NEW{Reconstruction time and novel view quality results for different methods. The first section compares our method with others that use unposed images, while the second section employs Structure from Motion (SfM). Reported runtimes include pose optimization. COLMAP intrinsics are used for both Photo-SLAM and MonoGS. The {\colorbox{firstcolor}{best}} and {\colorbox{secondcolor}{second best}} are color coded for pose-free methods.}}
\vspace{-7pt}
\begin{tabular}{l|cccc|cccc|cccc}
\toprule
 &
\multicolumn{4}{c|}{\textsc{TUM}} &
\multicolumn{4}{c|}{\textsc{MipNeRF360}} &
\multicolumn{4}{c}{\textsc{StaticHikes}}  \\
	{} & PSNR$^\uparrow$ & SSIM$^\uparrow$ & LPIPS$^\downarrow$ & Time$^\downarrow$ &
	PSNR$^\uparrow$ & SSIM$^\uparrow$ & LPIPS$^\downarrow$ & Time$^\downarrow$ &
	PSNR$^\uparrow$ & SSIM$^\uparrow$ & LPIPS$^\downarrow$ & Time$^\downarrow$ \\
\midrule
Photo-SLAM & 
\second{19.30} & \second{0.700} & 0.382 & \second{0:02:12} & \second{16.54} & \second{0.505} & \second{0.603} & \second{0:02:11} & 14.13 & \second{0.316} & 0.660 & \second{0:02:01} \\
MonoGS & 
16.60 & 0.682 & \second{0.381} & 0:16:18 & 14.46 & 0.436 & 0.663 & 0:04:05 & \second{15.46} & 0.301 & \second{0.659} & 0:09:19 \\
Ours & 
\first{23.02} & \first{0.821} & \first{0.250} & \first{0:00:50} & \first{24.31} & \first{0.775} & \first{0.300} & \first{0:01:02} & \first{20.40} & \first{0.589} & \first{0.365} & \first{0:01:30} \\
\hline
\hline
GLOMAP + Taming 3DGS (7k) & 
25.29 & 0.868 & 0.191 & 0:03:33 & 27.52 & 0.866 & 0.226 & 0:08:50 & 20.23 & 0.537 & 0.465 & 1:02:34 \\
COLMAP + 3DGS (7k)  & 
24.75 & 0.874 & 0.175 & 0:04:11 & 27.82 & 0.877 & 0.212 & 0:09:52 & 20.40 & 0.538 & 0.463 & 2:33:57 \\
COLMAP + 3DGS (30k) &
25.34 & 0.881 & 0.157 & 0:09:44 & 29.66 & 0.906 & 0.170 & 0:26:05 & 24.08 & 0.757 & 0.267 & 2:47:11 \\
\bottomrule
\end{tabular}
\end{table*}

\begin{table*}
\caption{\label{tab:slam-eval-low-res} \NEW{Novel view quality results for different methods that require low-resolution input. We use COLMAP intrinsics for COLMAP Free 3DGS.}}
\vspace{-7pt}
\begin{tabular}{l|cccc|cccc|cccc}
\toprule
 &
\multicolumn{4}{c|}{\textsc{TUM}} &
\multicolumn{4}{c|}{\textsc{MipNeRF360}} &
\multicolumn{4}{c}{\textsc{StaticHikes}}  \\
	{} & PSNR$^\uparrow$ & SSIM$^\uparrow$ & LPIPS$^\downarrow$ & Time$^\downarrow$ &
	PSNR$^\uparrow$ & SSIM$^\uparrow$ & LPIPS$^\downarrow$ & Time$^\downarrow$ &
	PSNR$^\uparrow$ & SSIM$^\uparrow$ & LPIPS$^\downarrow$ & Time$^\downarrow$ \\
\midrule
DROID-Splat &
\second{19.49} & \second{0.721} & \second{0.325} & \second{0:06:35} & \first{25.87} & \second{0.776} & \second{0.258} & \second{0:11:18} & \second{19.65} & \second{0.470} & \second{0.506} & \second{0:09:26} \\
CF-3DGS &
15.05 & 0.578 & 0.405 & 1:10:27 & 13.52 & 0.295 & 0.621 & 1:08:14 & 15.21 & 0.301 & 0.560 & 7:52:54 \\
Ours  &
\first{22.45} & \first{0.815} & \first{0.225} & \first{0:01:11} & \second{25.80} & \first{0.834} & \first{0.182} & \first{0:00:55} & \first{21.93} & \first{0.673} & \first{0.270} & \first{0:01:32} \\
\bottomrule
\end{tabular}
\end{table*}

\subsection{Methodology}
\label{sec:eval}

We present comparisons to two sets of methods. First we compare to state-of-the-art methods that do not require camera poses as input. These are mainly SLAM/3DGS and pose-free 3DGS solutions.
We selected methods for comparison based on code availability, reported performance, and the ability to handle as many scenes types as possible. 
For pose-free approaches that create a 3DGS scene, we compare to Photo-SLAM~\cite{hhuang2024photoslam}, DROID-Splat~\cite{homeyer2024droid} and MonoGS~\cite{gsslam2024}, and finally CF-3DGS \cite{Fu_2024_CVPR}, all presented in 2024. 

We also present two baselines. First, standard 3DGS (i.e., the release from the official github repo, using standard COLMAP parameters) for 7K and 30K iterations; the time reported is the \emph{total} of all the COLMAP processing and the 3DGS optimization. The second baseline uses Taming 3DGS~\cite{taming3dgs} which is the fastest current 3DGS optimization, coupled with a best effort approach to accelerate SfM pose estimation using GLOMAP~\cite{pan2024glomap}. Specifically we run the COLMAP feature extractor, sequential matcher and then the GLOMAP mapper to get poses and SfM points.
This latter baseline can be considered the current fastest best-practice, non-incremental solution to pose estimation and 3DGS optimization. 
\NEW{Since the \textsc{TUM} dataset features denser capture, we run methods without keyframe selection (Taming 3DGS, 3DGS, and COLMAP Free 3DGS) on fewer images. Specifically, we select every 3rd, 15th, and 10th frame for fr1, fr2, and fr3, respectively. This approach aligns the total number of frames more closely with our number of registered keyframes while retaining all images from the test set. Note that the reported time for our method includes automatic keyframe selection.}

DROID-Splat and CF-3DGS cannot handle full resolution images. As a result, we present a separate table (Tab.~\ref{tab:slam-eval-low-res}), with these methods at a resolution where they work both as well as possible for each dataset (446x336 for \textsc{TUM} and 640 width for \textsc{MipNeRF360} and \textsc{StaticHikes}). \NEW{Our method requires higher resolution inputs, as XFeat~\cite{potje2024xfeat} performs optimally within the 1 to 2 megapixel range. For comparisons, we upsample the resized images by a factor of two in both dimensions before processing them with our method. Metrics are then reported on the appropriately downsampled images.}
Another issue is the specification of the set of test images for evaluation. Different approaches are used for different methods, since in some cases not all images have an estimated pose, and thus the test set is often different for each method, even for the same scene. 
\NEW{We defined a single evaluation protocol, using every $n^\text{th}$ image as a test view, where $n$ is 8 and 10 for \textsc{MipNeRF360} and \textsc{StaticHikes}, as proposed by their authors, and 30 for \textsc{TUM}, as the baseline between frames is small.} This required specific modifications for each method (please see the Appendix).

The second comparison is for methods treating large-scale scenes that cannot be handled by standard 3DGS. Specifically, we compare to H3DGS~\cite{hierarchicalgaussians24}. 
For this comparison, we use the  front camera from the \textsc{SmallCity} and \textsc{Wayve} datasets as well as our \textsc{CityWalk} dataset. We provide the COLMAP calibration using the method from H3DGS for all these scenes, since it is one of the few ways to get camera poses with SfM for scenes of this size.
We also evaluate pose estimation quality for the test views, comparing to ground truth poses when these exist (i.e., \textsc{TUM}) and using the COLMAP poses as ``pseudo-ground truth'' following standard practice and using the RMSE APE and RPE metrics~\cite{ATE}.

\begin{table*}%
\NEW{
\caption{\label{tab:refine} \NEW{Results with additional optimization. Our method achieves quality similar to Taming 3DGS (7k).}}
\vspace{-7pt}
\begin{tabular}{l|cccc|cccc|cccc}
\toprule
 &
\multicolumn{4}{c|}{\textsc{TUM}} &
\multicolumn{4}{c|}{\textsc{MipNeRF360}} &
\multicolumn{4}{c}{\textsc{StaticHikes}}  \\
	{\# epochs} & PSNR$^\uparrow$ & SSIM$^\uparrow$ & LPIPS$^\downarrow$ & Time$^\downarrow$ &
	PSNR$^\uparrow$ & SSIM$^\uparrow$ & LPIPS$^\downarrow$ & Time$^\downarrow$ &
	PSNR$^\uparrow$ & SSIM$^\uparrow$ & LPIPS$^\downarrow$ & Time$^\downarrow$ \\
\midrule
10 &
24.38 & 0.843 & 0.224 & \first{0:00:58} & 25.88 & 0.815 & 0.265 & \first{0:01:20} & 21.06 & 0.620 & 0.338 & \first{0:02:00} \\
25 &
25.09 & 0.853 & 0.210 & \second{0:01:09} & 26.51 & 0.830 & 0.247 & \second{0:01:47} & 21.26 & 0.635 & 0.322 & \second{0:02:42} \\
50 &
\second{25.63} & 0.862 & 0.198 & 0:01:28 & 27.08 & 0.842 & 0.233 & 0:02:34 & \second{21.43} & \second{0.649} & \second{0.308} & 0:03:50 \\
100 &
\first{26.08} & \second{0.866} & \first{0.189} & 0:02:05 & \second{27.17} & \second{0.848} & \first{0.224} & 0:04:09 & \first{21.57} & \first{0.660} & \first{0.296} & 0:06:08 \\
\midrule
Taming 3DGS (7k) &
25.29 & \first{0.868} & \second{0.191} & 0:03:33 & \first{27.52} & \first{0.866} & \second{0.226} & 0:08:50 & 20.23 & 0.537 & 0.465 & 1:02:34 \\
\bottomrule
\end{tabular}
}
\end{table*}

\subsection{Novel View Synthesis Quality}
\label{sec:nvs-eval}

In Tab.~\ref{tab:slam-eval},~\ref{tab:slam-eval-low-res}, we show the average results for each method broken down by dataset type for the SLAM set of methods. For each method we show the standard metrics PSNR, SSIM and LPIPS as well as the average time for full processing, i.e., the total time for pose estimation and 3DGS optimization. 
\NEW{The average times for GLOMAP pose optimization are 0:02:02, 0:07:17, and 1:00:26 for \textsc{TUM}, \textsc{MipNeRF360}, and \textsc{StaticHikes}, respectively.
For \textsc{TUM}, we use a subset of the images for CF-3DGS and the SfM methods, as they do not feature keyframing, whereas other methods process all images. 
Additionally, SfM-based approaches require the entire dataset before processing, as their mappers reorder the images. This prevents live feedback and obtaining the reconstruction immediately at the end of the capture.
}

\begin{figure*}
\setlength{\tabcolsep}{1pt}
\renewcommand{\arraystretch}{0.6}
\begin{tabular}{cccccccc}
	{} &	{Taming 3DGS}  & {Photo-SLAM} & {MonoGS} & {Ours} & {GT} \\
	\centeredtab{\rot{\textsc{TUM}}} &  
	\picwithtext{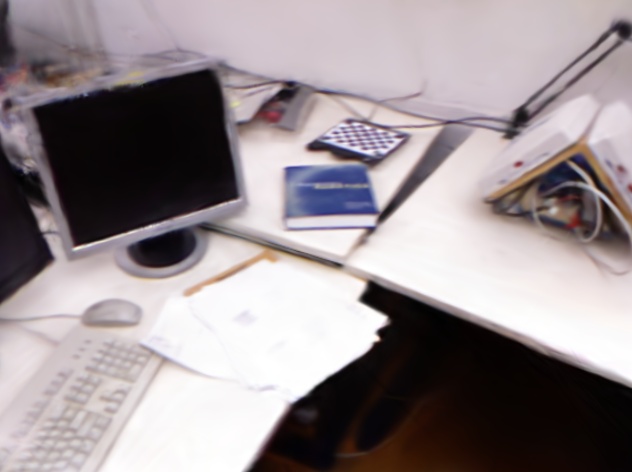}{.193\linewidth}{0:02:20, 23.74 dB} &
	\picwithtext{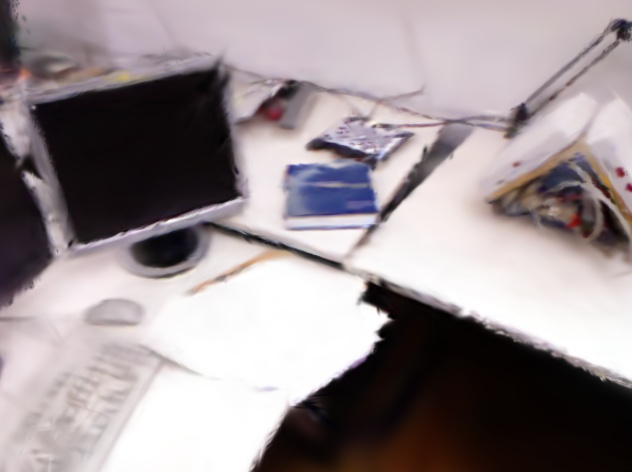}{.193\linewidth}{0:00:35, 18.50 dB} &
	\picwithtext{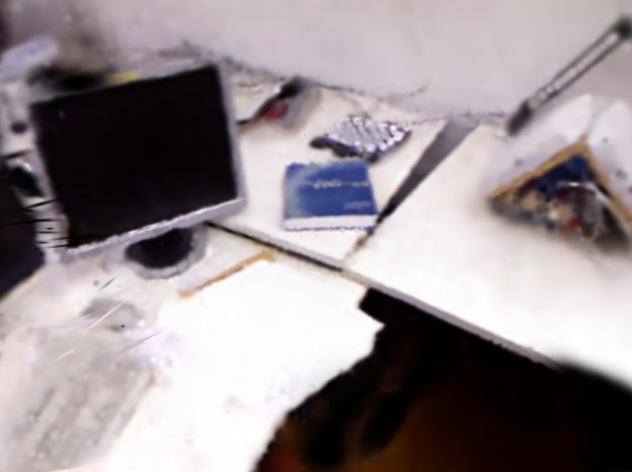}{.193\linewidth}{0:08:45, 15.02 dB} & 
	\picwithtext{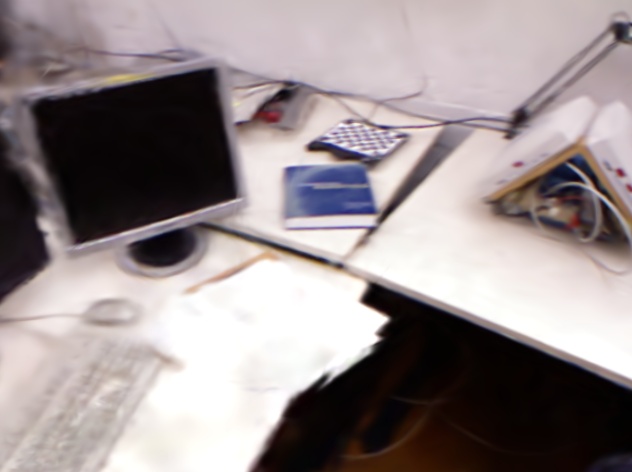}{.193\linewidth}{0:00:28, 20.36 dB} &
	\centeredtab{\includegraphics[width=.193\linewidth]{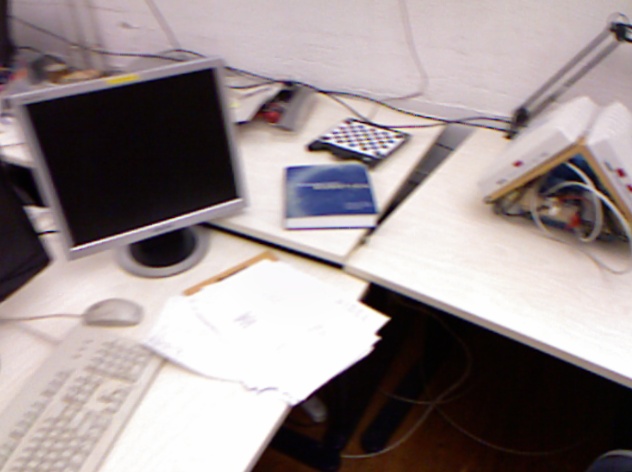}} \\
	\centeredtab{\rot{\textsc{MipNeRF360}}} &
	\picwithtext{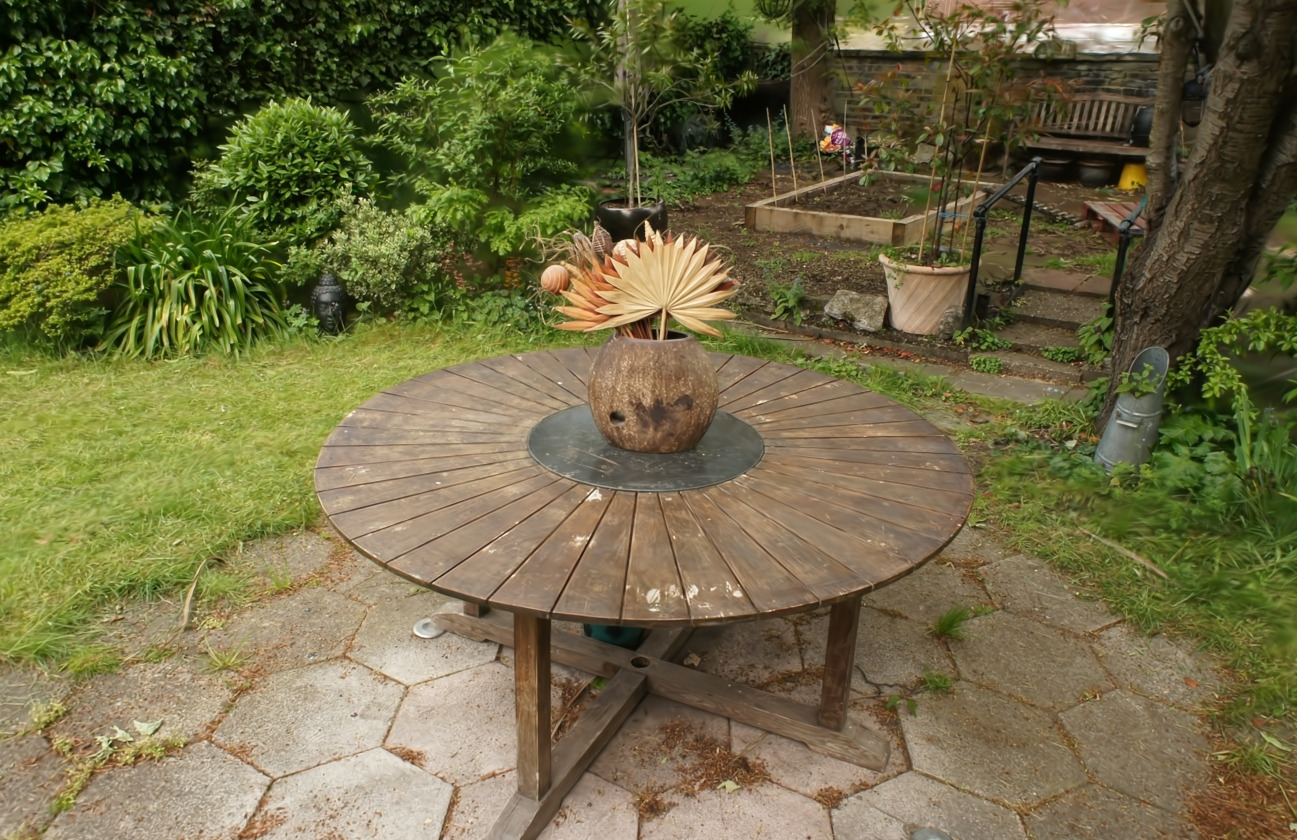}{.193\linewidth}{0:07:48, 25.78 dB}  &
	\picwithtext{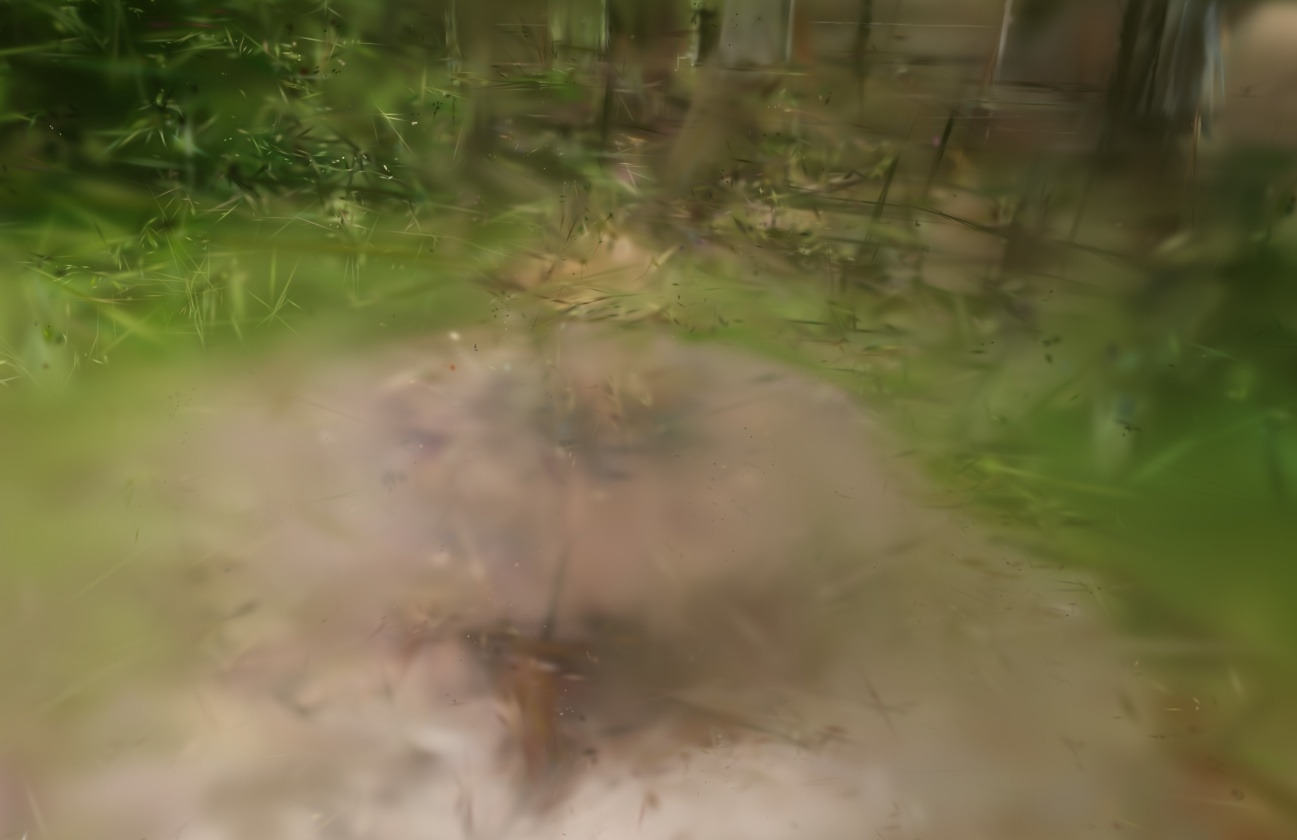}{.193\linewidth}{0:01:42, 16.02 dB} &
	\picwithtext{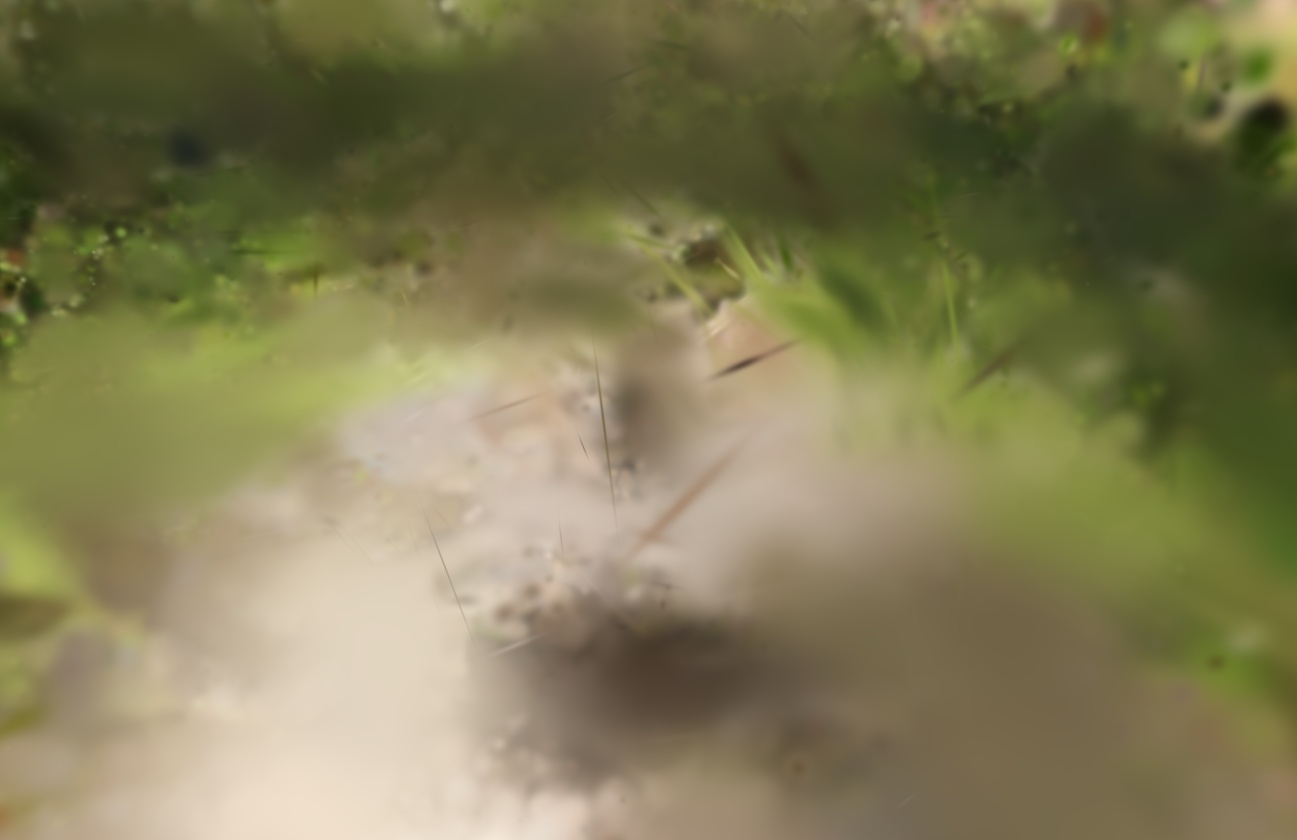}{.193\linewidth}{0:08:21, 14.43 dB} &
	\picwithtext{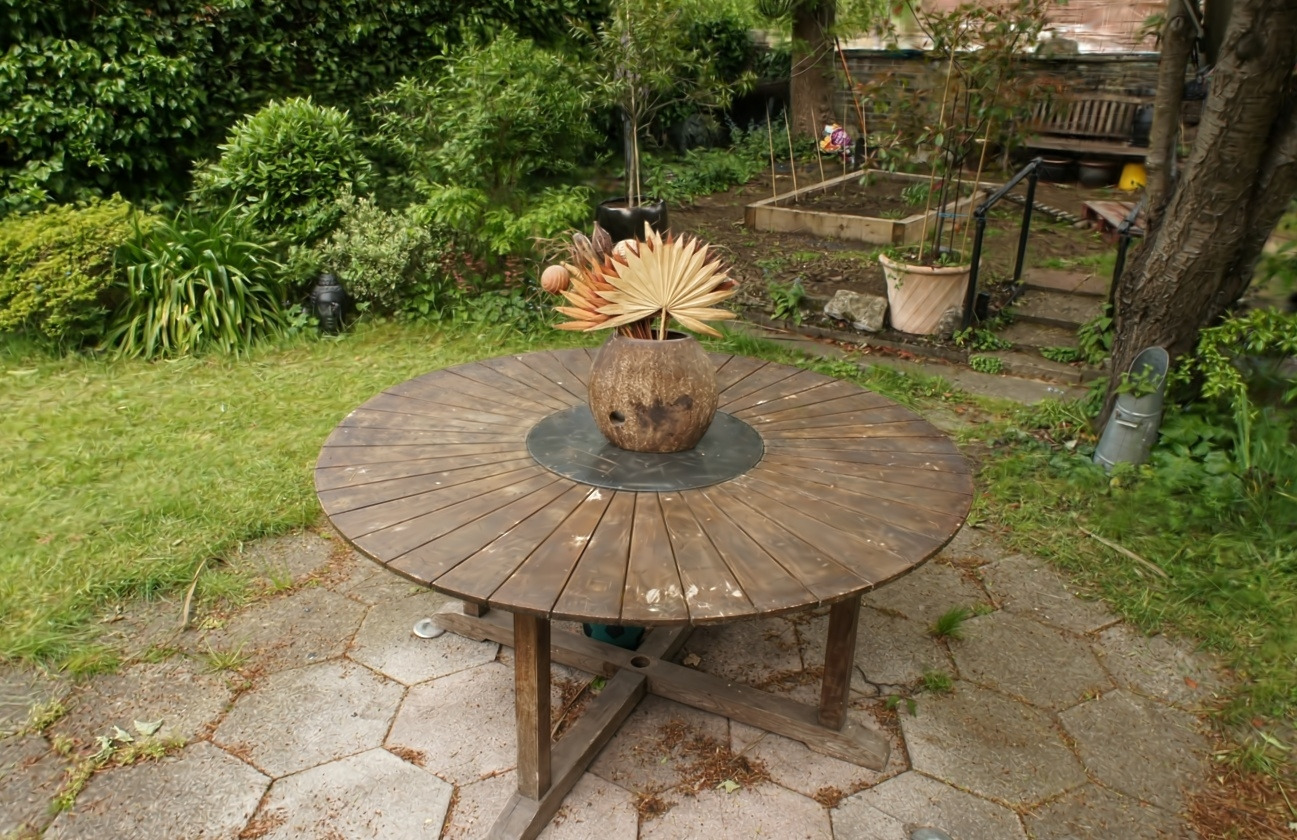}{.193\linewidth}{0:00:55, 24.51 dB} &
	\centeredtab{\includegraphics[width=.193\linewidth]{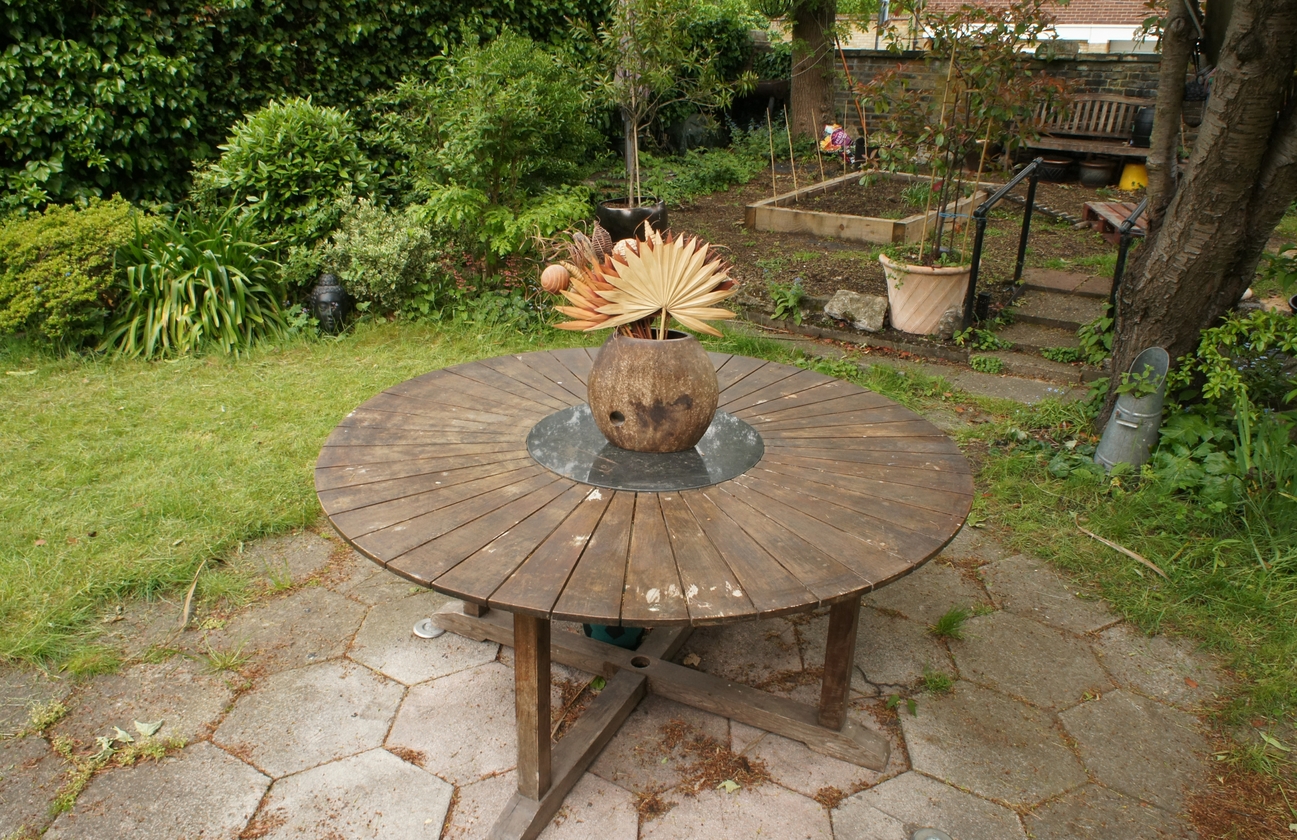}} \\
	\centeredtab{\rot{\textsc{StaticHikes}}} & 
	\picwithtext{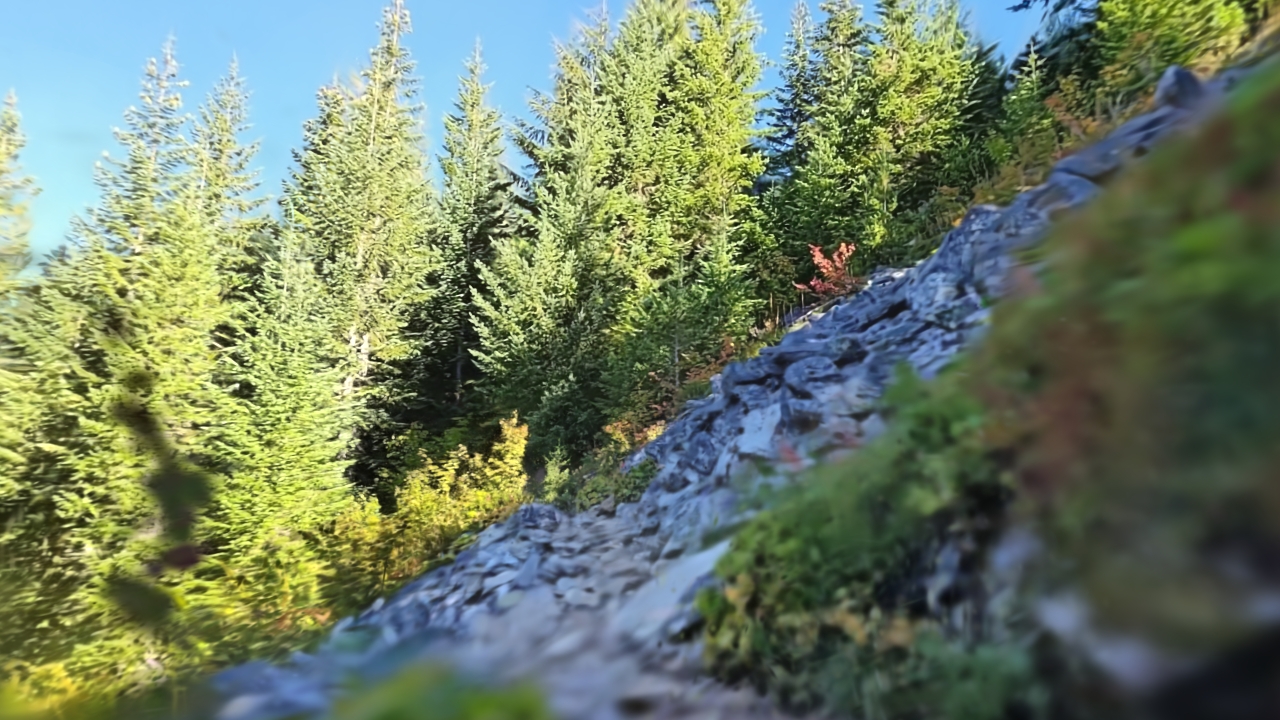}{.193\linewidth}{0:43:17, 17.70 dB} &
	\picwithtext{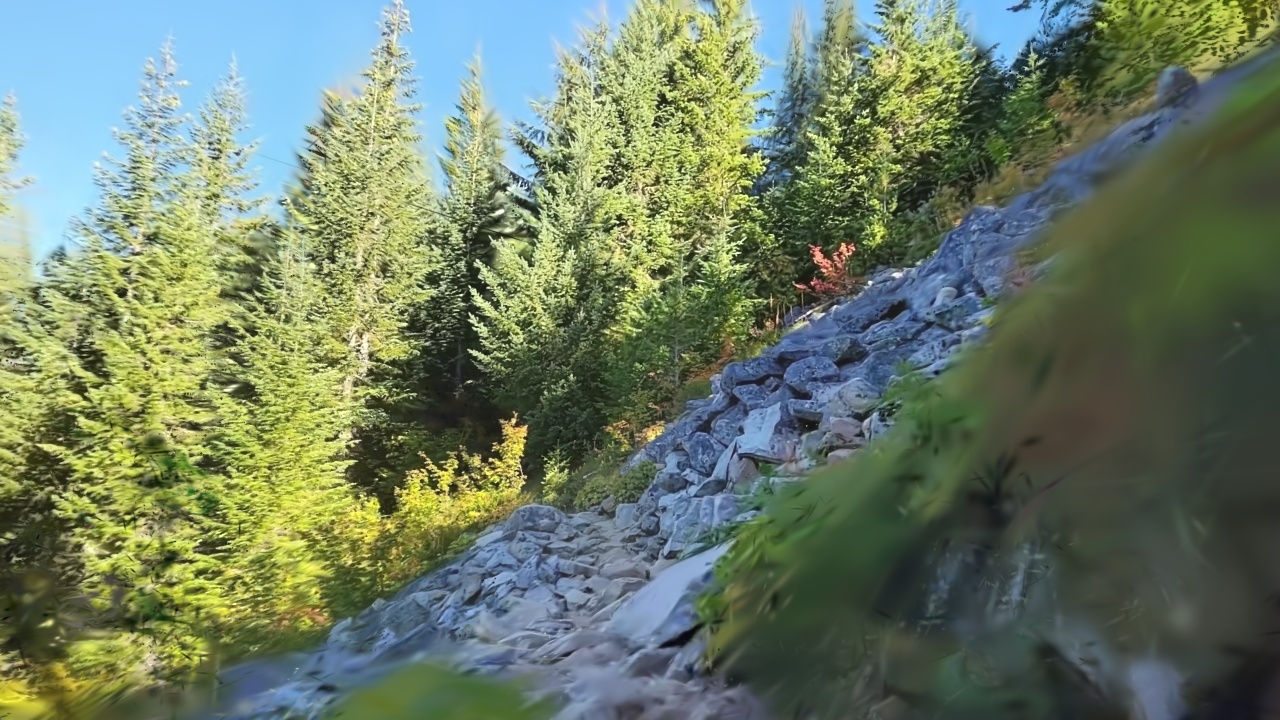}{.193\linewidth}{0:02:28, 14.68 dB} & 
	\picwithtext{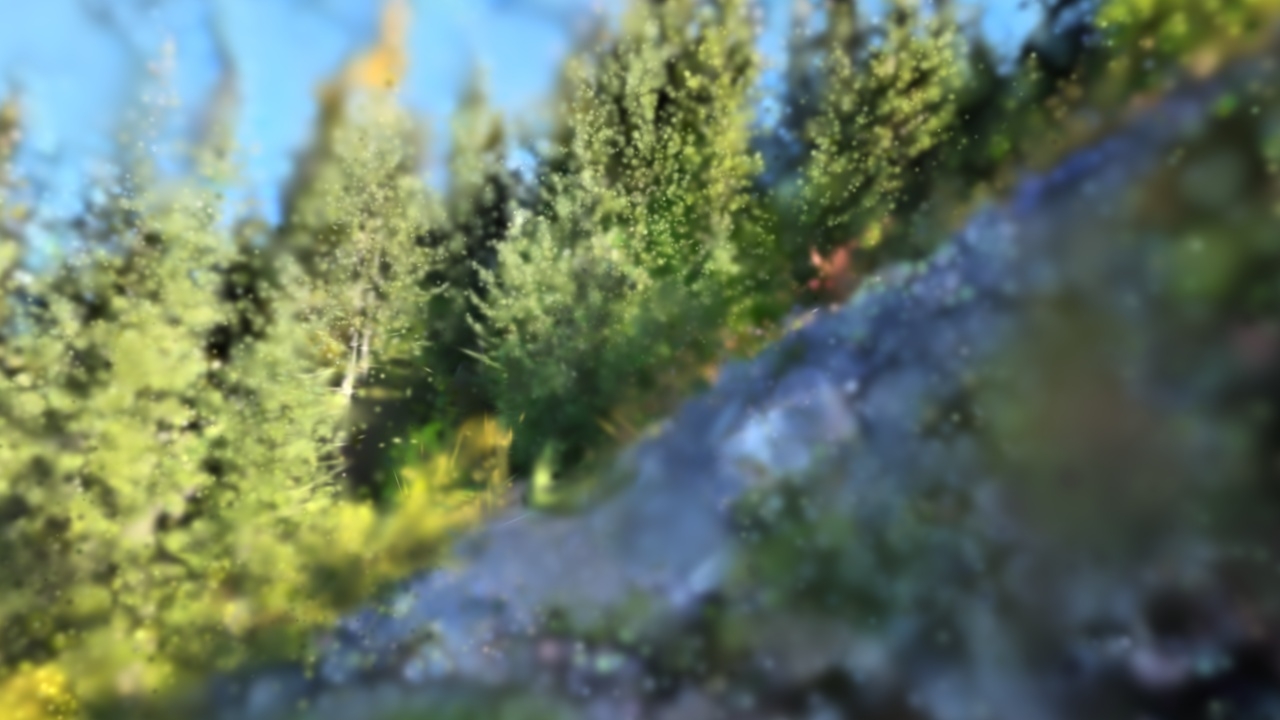}{.193\linewidth}{0:17:52, 13.18 dB} & 
	\picwithtext{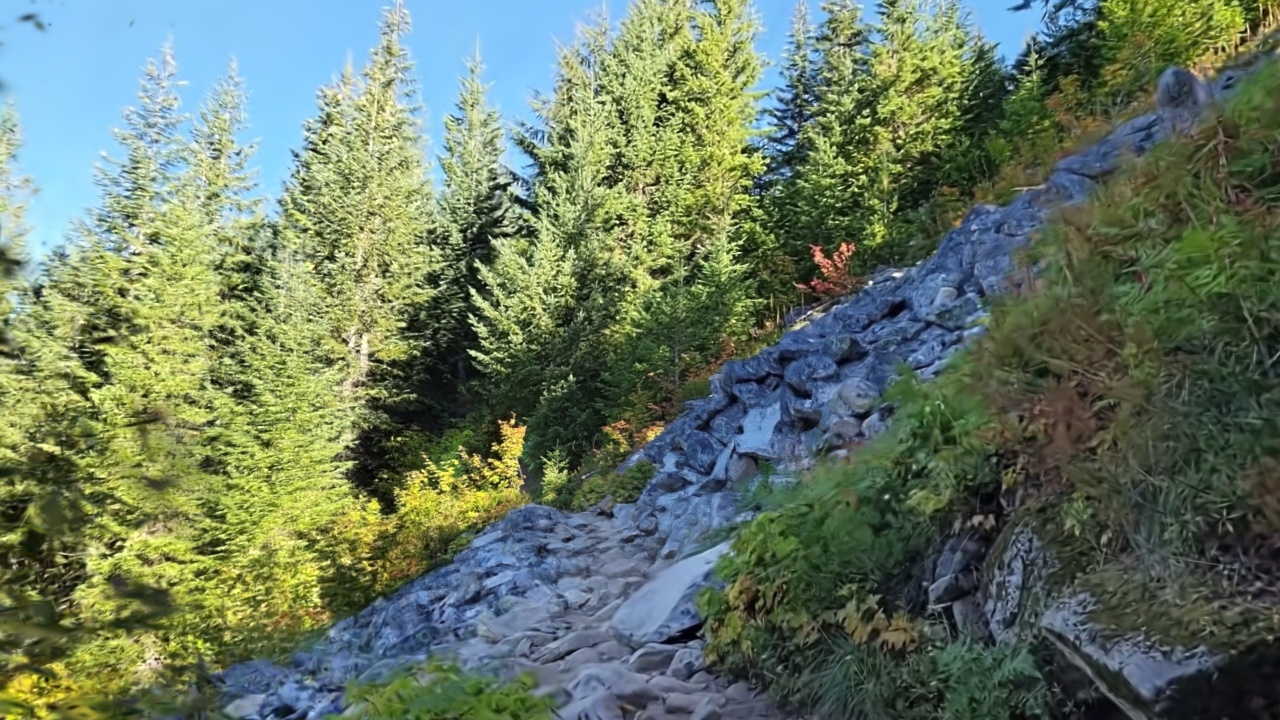}{.193\linewidth}{0:01:54, 18.10 dB} & 
	\centeredtab{\includegraphics[width=.193\linewidth]{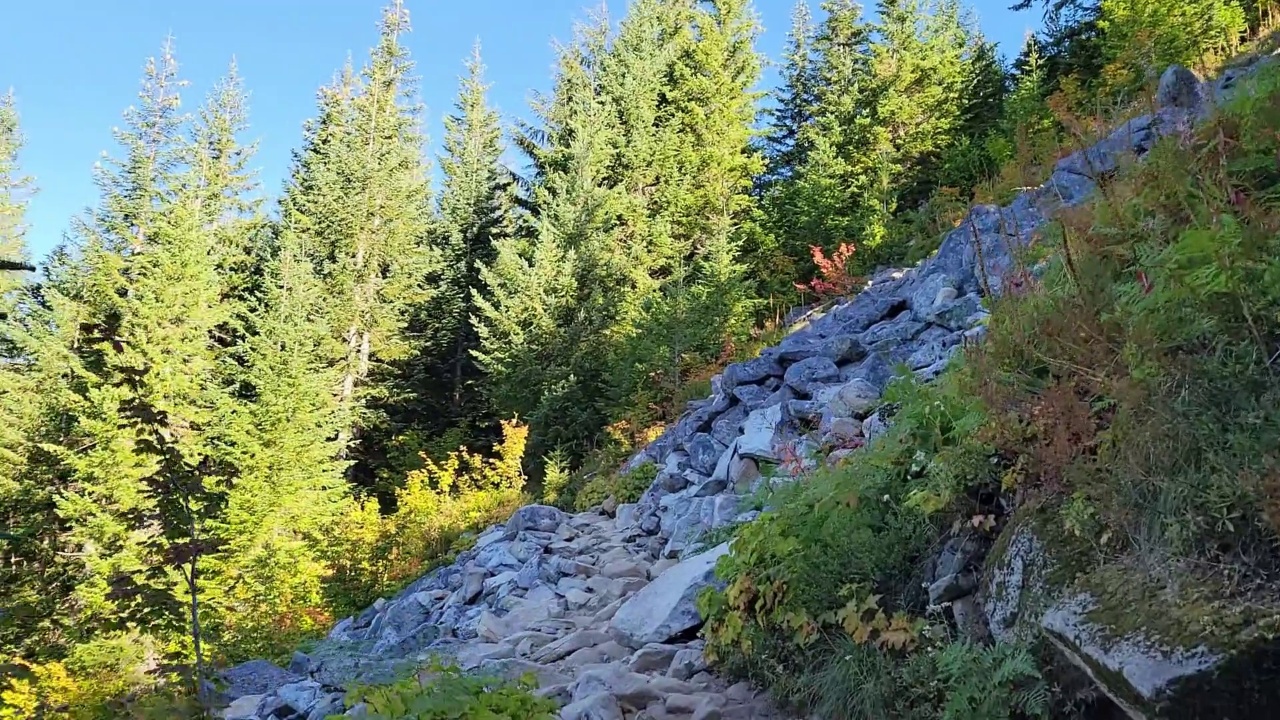}} \\
\end{tabular}
\vspace{-7pt}
\caption{
\label{fig:compare-slam}
	Qualitative comparison for the three datasets used, for Taming 3DGS, Photo-Slam, MonoGS. We include the images of these methods for the test views in Fig.~\ref{fig:compare-slam2} in supplemental. 
	\NEW{We show the scene reconstruction time and PSNR.}
	Note that Taming 3DGS requires significantly more time since it uses offline SfM for camera pose estimation. Our approach provides better visual quality for these scenes compared to other methods that take unposed images as input.
}
\end{figure*}

\begin{figure*}
    \setlength{\tabcolsep}{1pt}
    \renewcommand{\arraystretch}{0.6}
\begin{tabular}{cccccccc}
{}&	{CF-3DGS} & {DROID-Splat} & {Ours} & {GT} \\
	\centeredtab{\rot{\textsc{TUM}}} & 
	\picwithtext{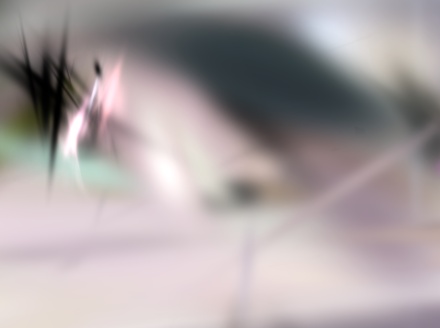}{.242\linewidth}{0:54:34, 15.89 dB} & 
	\picwithtext{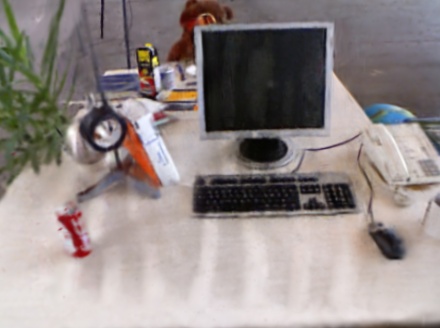}{.242\linewidth}{0:08:31, 15.51 dB} & 
	\picwithtext{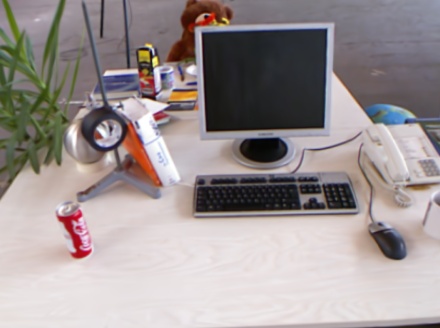}{.242\linewidth}{0:01:02, 25.80 dB} & 
	\centeredtab{\includegraphics[width=.242\linewidth]{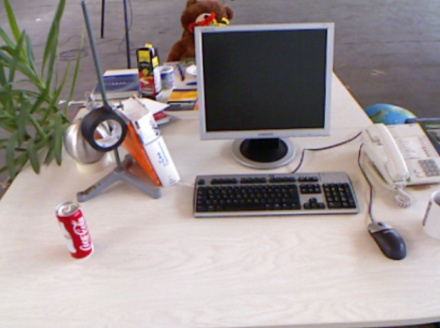}} \\
	\centeredtab{\rot{\textsc{MipNeRF360}}} & 
	\picwithtext{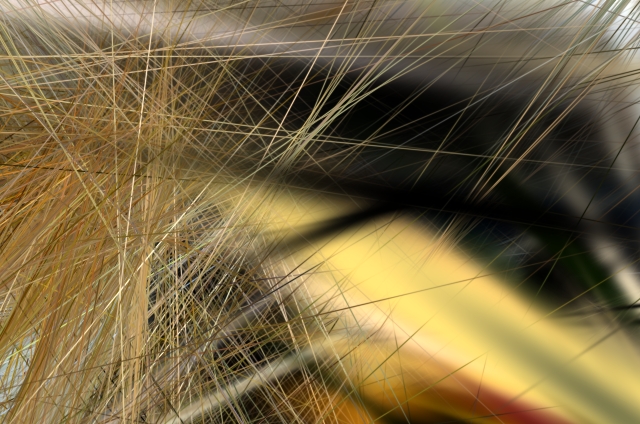}{.242\linewidth}{0:14:57, 11.98 dB} & 
	\picwithtext{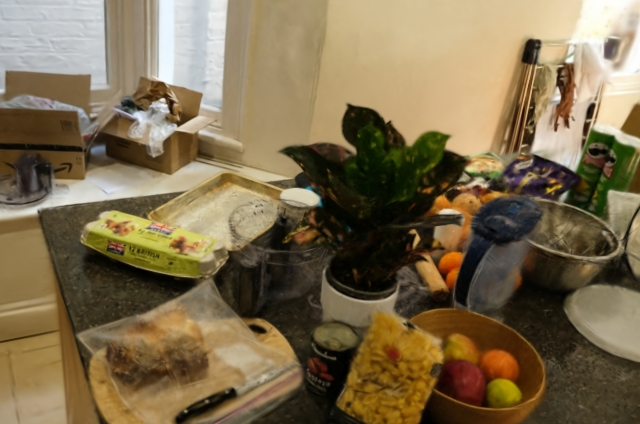}{.242\linewidth}{0:11:27, 25.94 dB} & 
	\picwithtext{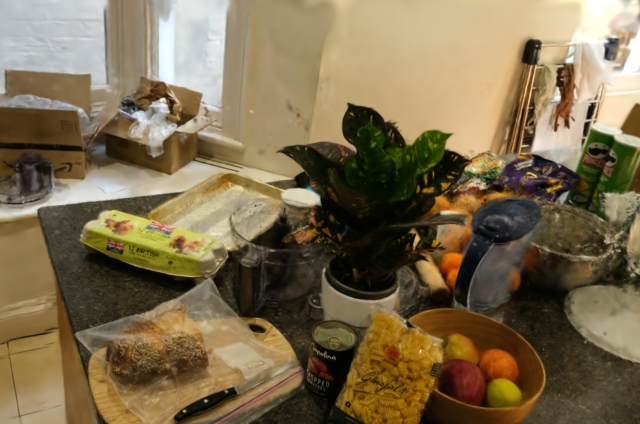}{.242\linewidth}{0:00:53, 24.94 dB} & 
	\centeredtab{\includegraphics[width=.242\linewidth]{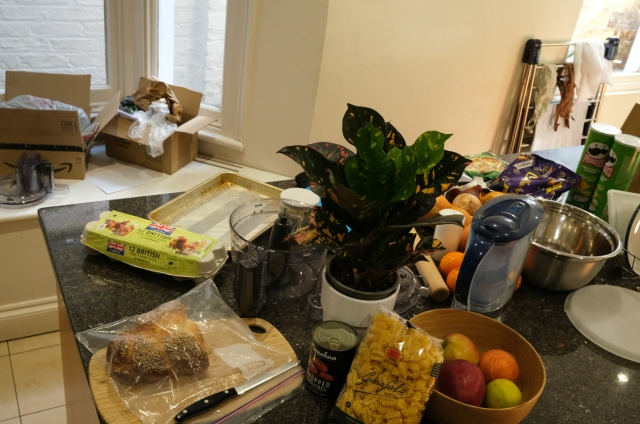}} \\
	\centeredtab{\rot{\textsc{StaticHikes}}} & 
	\picwithtext{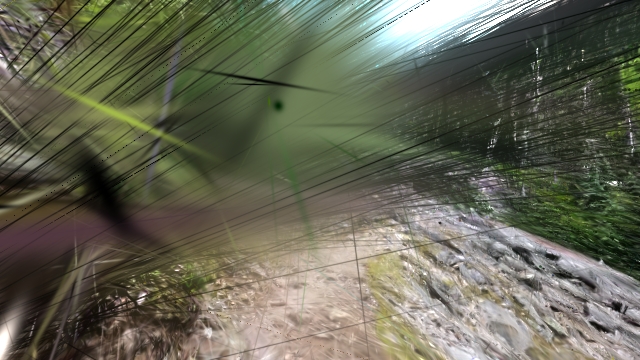}{.242\linewidth}{8:33:46, 17.80 dB} & 
	\picwithtext{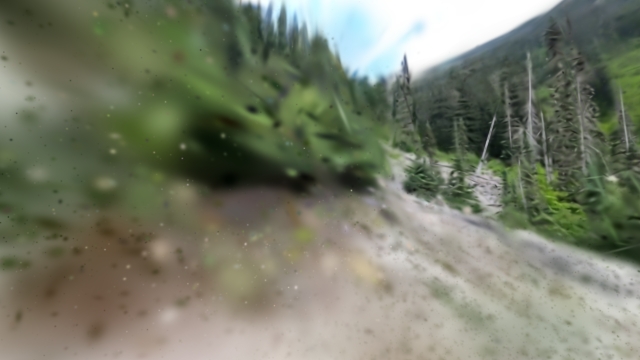}{.242\linewidth}{0:07:29, 17.46 dB} & 
	\picwithtext{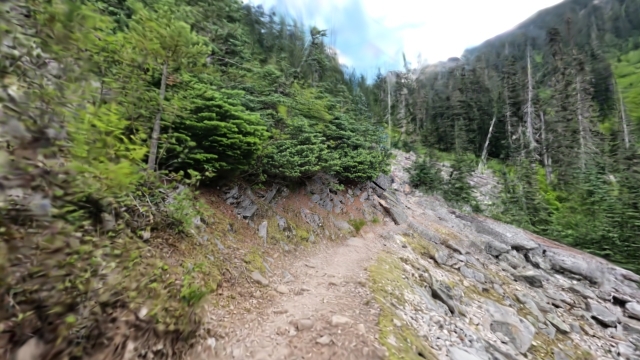}{.242\linewidth}{0:01:13, 23.72 dB} & 
	\centeredtab{\includegraphics[width=.242\linewidth]{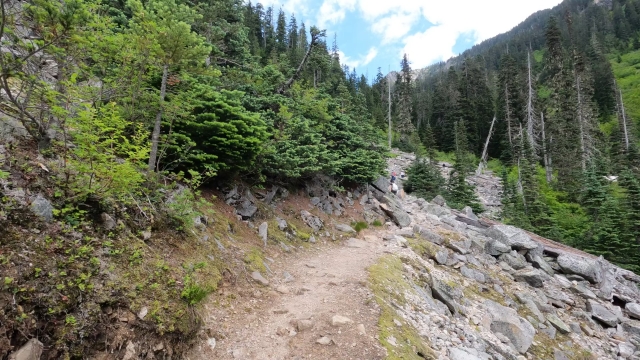}} \\
\end{tabular}
\vspace{-7pt}
\caption{
\label{fig:compare-slam2}
	Qualitative comparison of pose-free methods for the three datasets used, for CF-3DGS and DROID-Splat that only handle low resolution. We include the images of these methods for the test views in Fig.~\ref{fig:compare-slam} in supplemental. 
	\NEW{We show the scene reconstruction time and PSNR.}
We see that our method is competitve to previous approaches, and is robust to different capture styles.
}
\end{figure*}

We also show qualitative results visually comparing the different methods in Figures~\ref{fig:compare-slam} and~\ref{fig:compare-slam2}.
We see that the visual quality of our solution is on par or better than all competitors, for all types of scenes. DROID-Splat has good visual quality; our method tends to be sharper, but can have slightly lower fidelity.
SLAM methods perform well on the dense captures they are designed for, but visual quality can degrade or the method can even fail as the camera baseline becomes wider. 
Taming 3DGS and standard 3DGS have better visual quality, and work well for all scenes, but with the \NEW{high} computational overhead discussed previously, making them unsuitable for our on-the-fly reconstruction scenario.

After the training has processed all views, we can fine-tune Gaussians and cameras using the identified keyframes. To do this, we load anchors one-by-one. For each anchor, the associated cameras and Gaussians parameters are optimized further by randomly sampling all of the anchor's keyframes. Since only 3DGS optimization is performed, this process is fast enough, and we can repeat it for multiple \emph{epochs} to find an ideal overhead/quality tradefoff. 
\NEW{Tab.~\ref{tab:refine} shows that we reach Taming 3DGS (7k)'s quality.}
However, achieving quality beyond this requires a more involved solution; we discuss this as future work Sec.~\ref{sec:discuss}.

\begin{table*}
\caption{\label{tab:large-scale} Results for large scale scenes. For other methods time includes total time for COLMAP using the H3DGS process and the actual 3DGS optimization of each method.}
\vspace{-7pt}
\begin{tabular}{l|cccc|cccc|cccc}
\toprule
 &
\multicolumn{4}{c|}{\textsc{SmallCity*}} &
\multicolumn{4}{c|}{\textsc{Wayve*}} &
\multicolumn{4}{c}{\textsc{CityWalk}}  \\
  & PSNR$^\uparrow$ & SSIM$^\uparrow$ & LPIPS$^\downarrow$ & Time$^\downarrow$ &
	PSNR$^\uparrow$ & SSIM$^\uparrow$ & LPIPS$^\downarrow$ & Time$^\downarrow$ &
	PSNR$^\uparrow$ & SSIM$^\uparrow$ & LPIPS$^\downarrow$ & Time$^\downarrow$ \\
\midrule
{H3DGS} & 
21.17 & 0.679 & 0.285 & 2:55:28 & 20.80 & 0.737 & 0.227 & 7:29:45 & 11.78 & 0.557 & 0.560 & 22:09:20 \\
{Ours} & 
23.59 & 0.789 & 0.323 & 0:01:45 & 20.29 & 0.739 & 0.303 & 0:04:29 & 21.71 & 0.712 & 0.395 & 00:25:03 \\
\bottomrule
\end{tabular}
\end{table*}

We next show results for the large-scale scenes, where we compare to H3DGS. In Tab.~\ref{tab:large-scale} we see that the overhead of camera calibration using SfM approaches grows significantly with the scale of the scene. 
Capturing the scene \textsc{CityWalk} took 30min which is more than the 25min our method requires to process the scene; using H3DGS (one of the few methods that can handle captures this big) requires 22 hours of processing after capture is finished. In addition, the quality of the pose estimation is very low, leading to failure for novel view synthesis in several segments of the path.

\begin{figure}
    \setlength{\tabcolsep}{1pt}
    \renewcommand{\arraystretch}{0.6}
\begin{tabular}{ccc}
 {} & H3DGS & Ours \\
\centeredtab{\rot{\textsc{SmallCity*}}} & 
\picwithtext{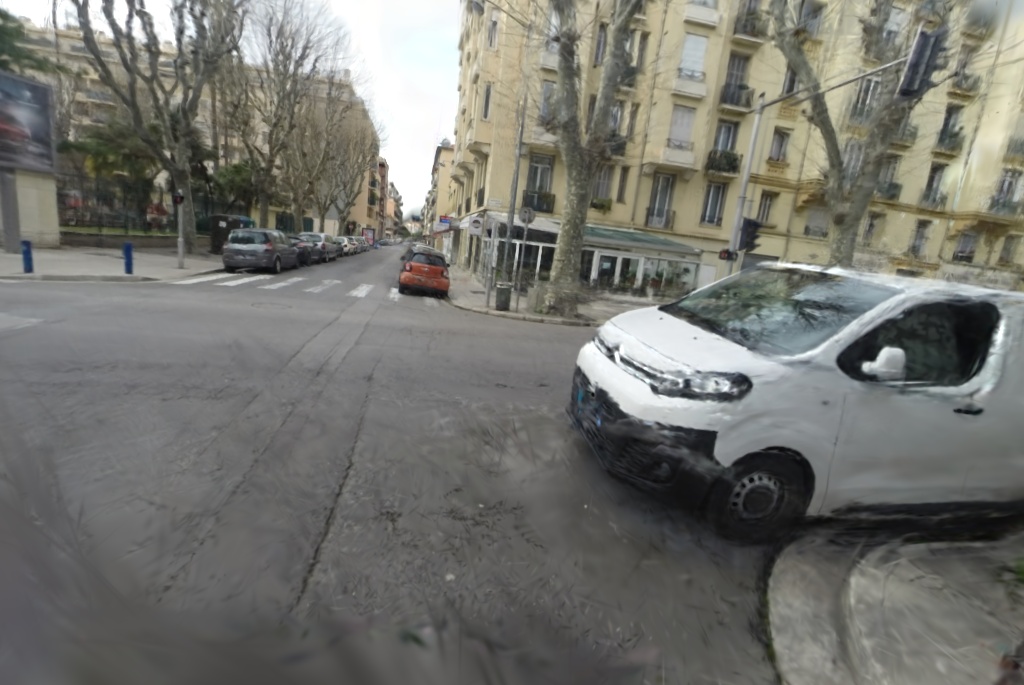}{.475\linewidth}{2:55:28, 21.17 dB} & 
\picwithtext{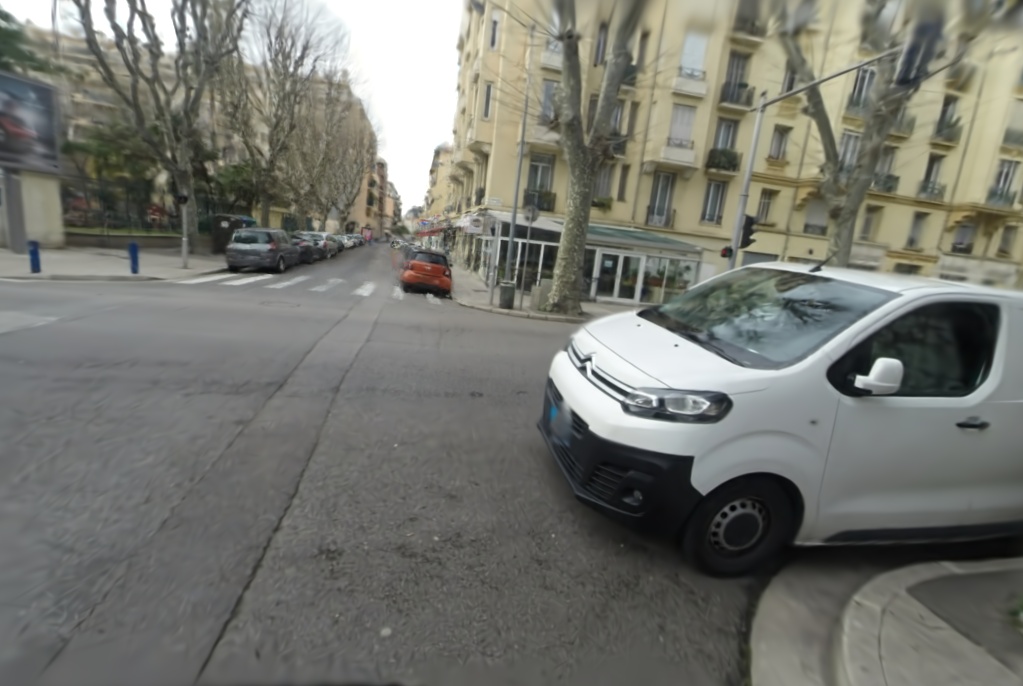}{.475\linewidth}{0:01:45, 23.59 dB} \\
\centeredtab{\rot{\textsc{Wayve*}}} & 
\picwithtext{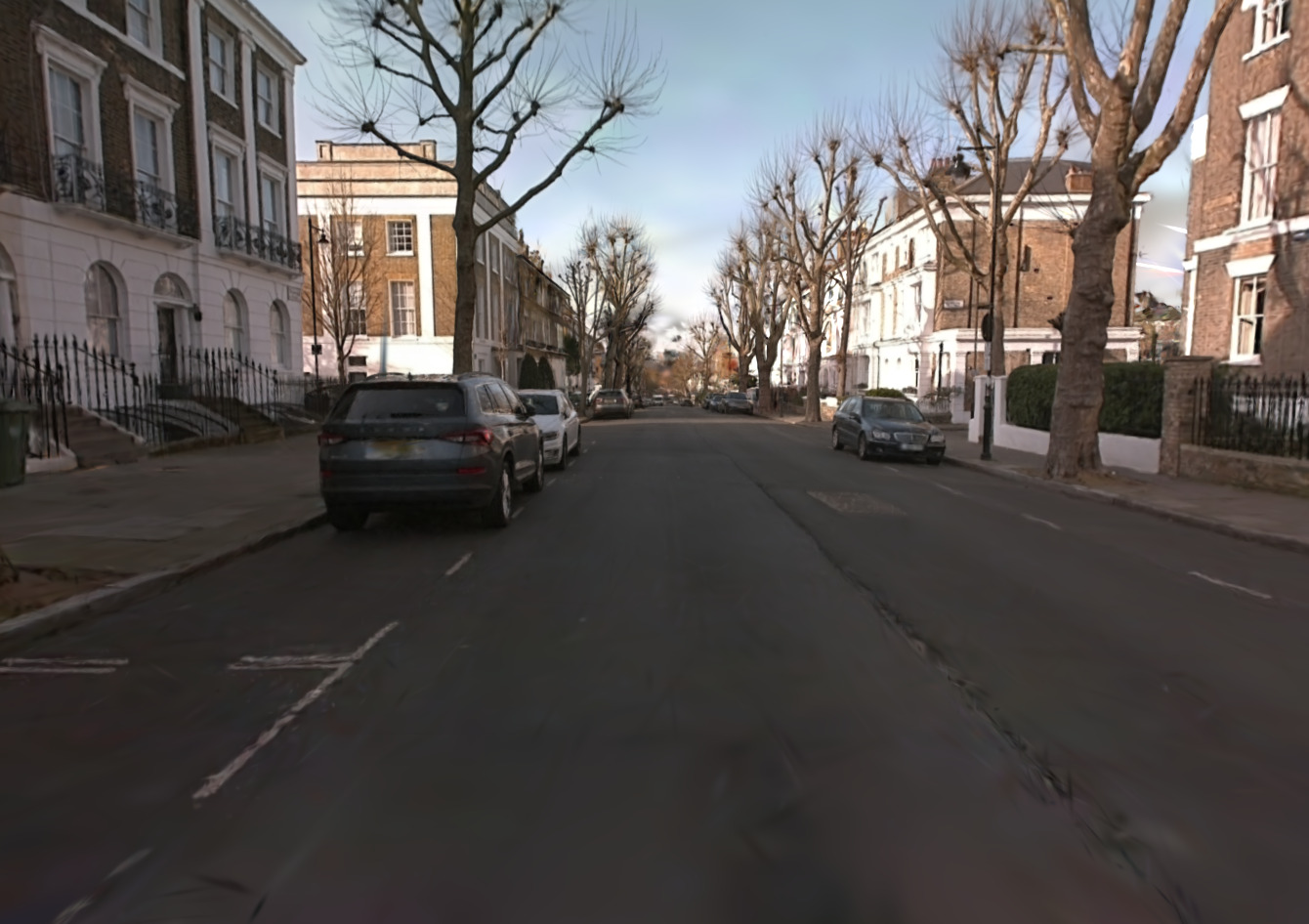}{.475\linewidth}{7:29:45, 20.80 dB} & 
\picwithtext{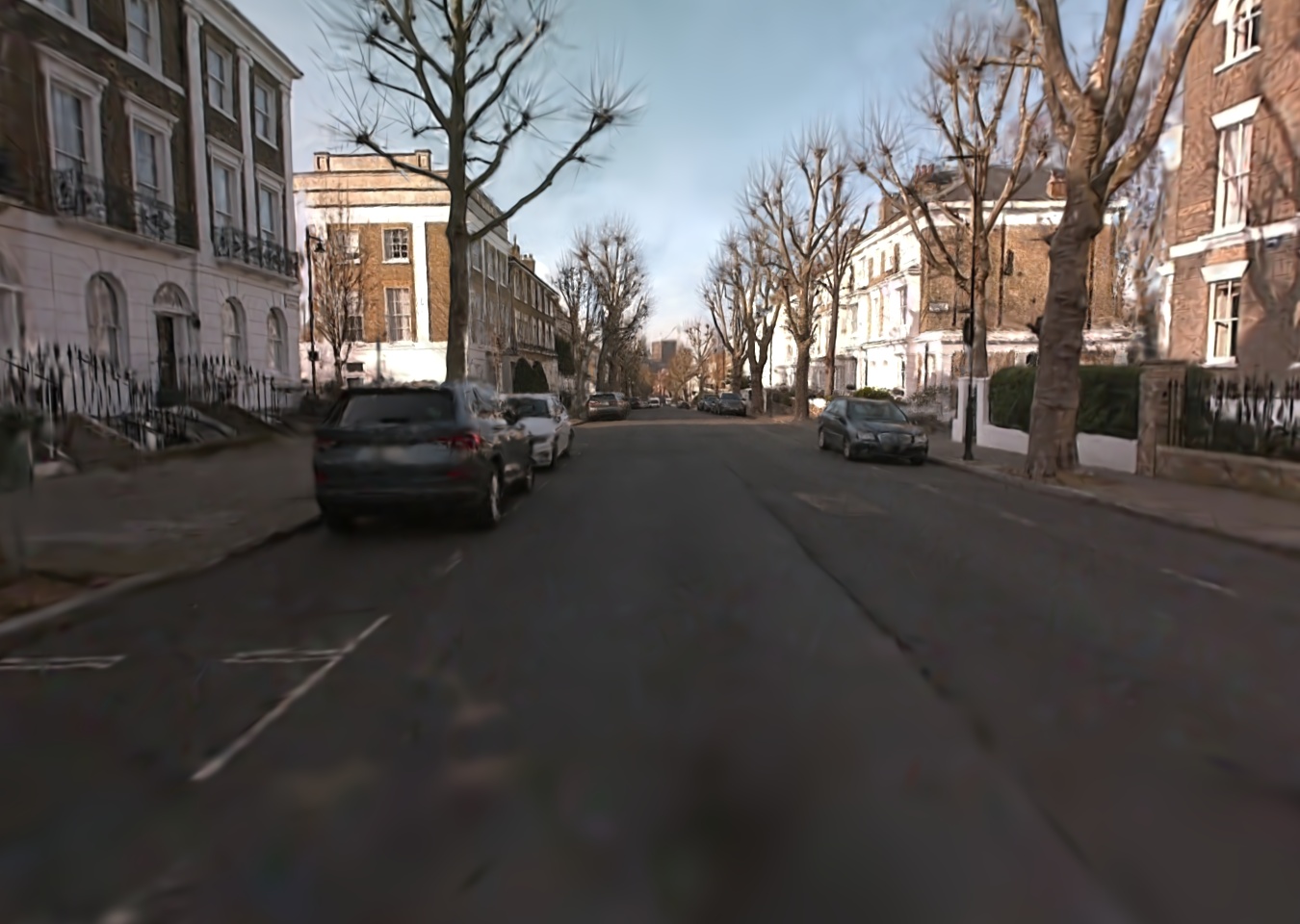}{.475\linewidth}{0:04:29, 20.29 dB} \\
\centeredtab{\rot{\textsc{CityWalk}}} & 
\picwithtext{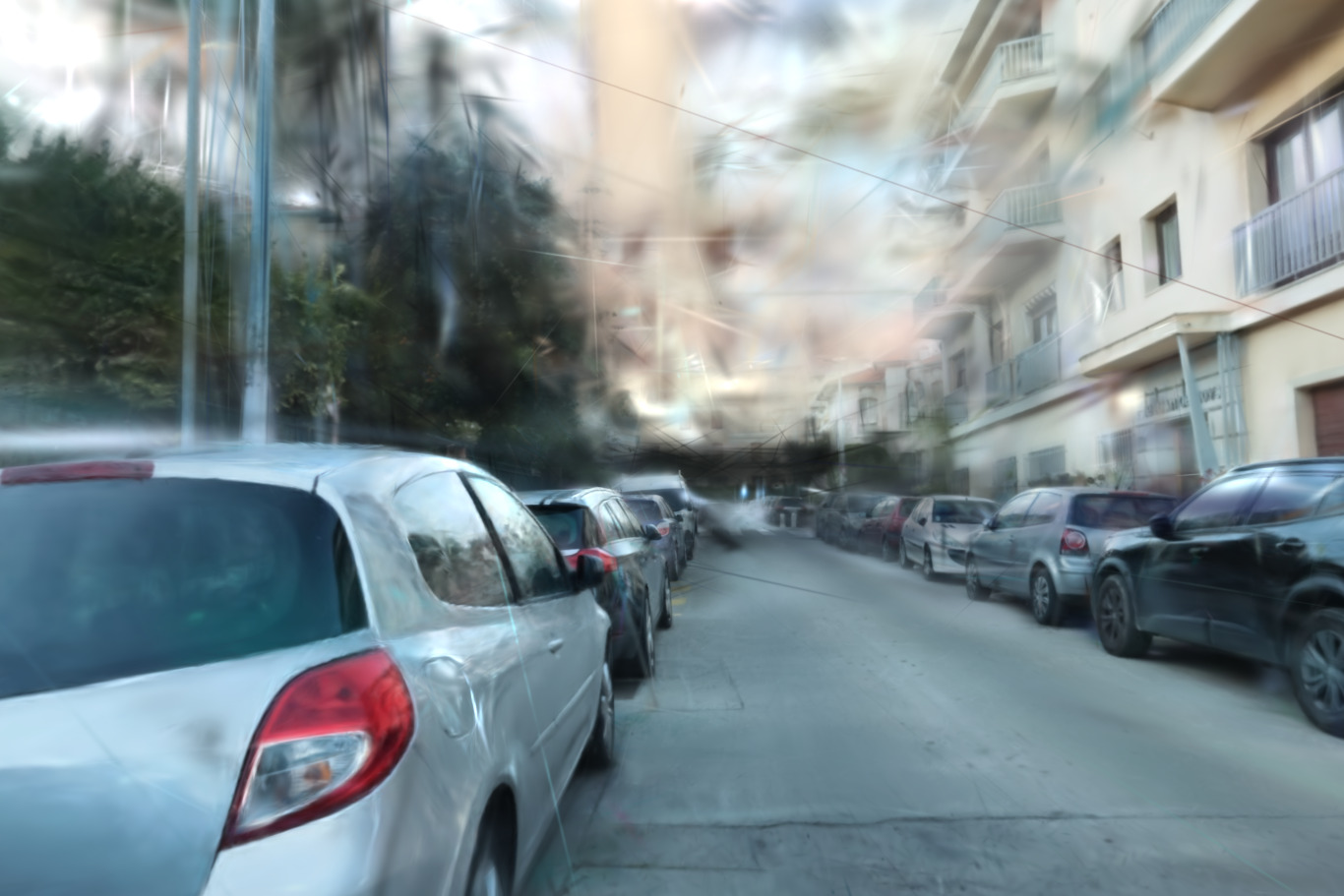}{.475\linewidth}{22:09:20, 11.78 dB} & 
\picwithtext{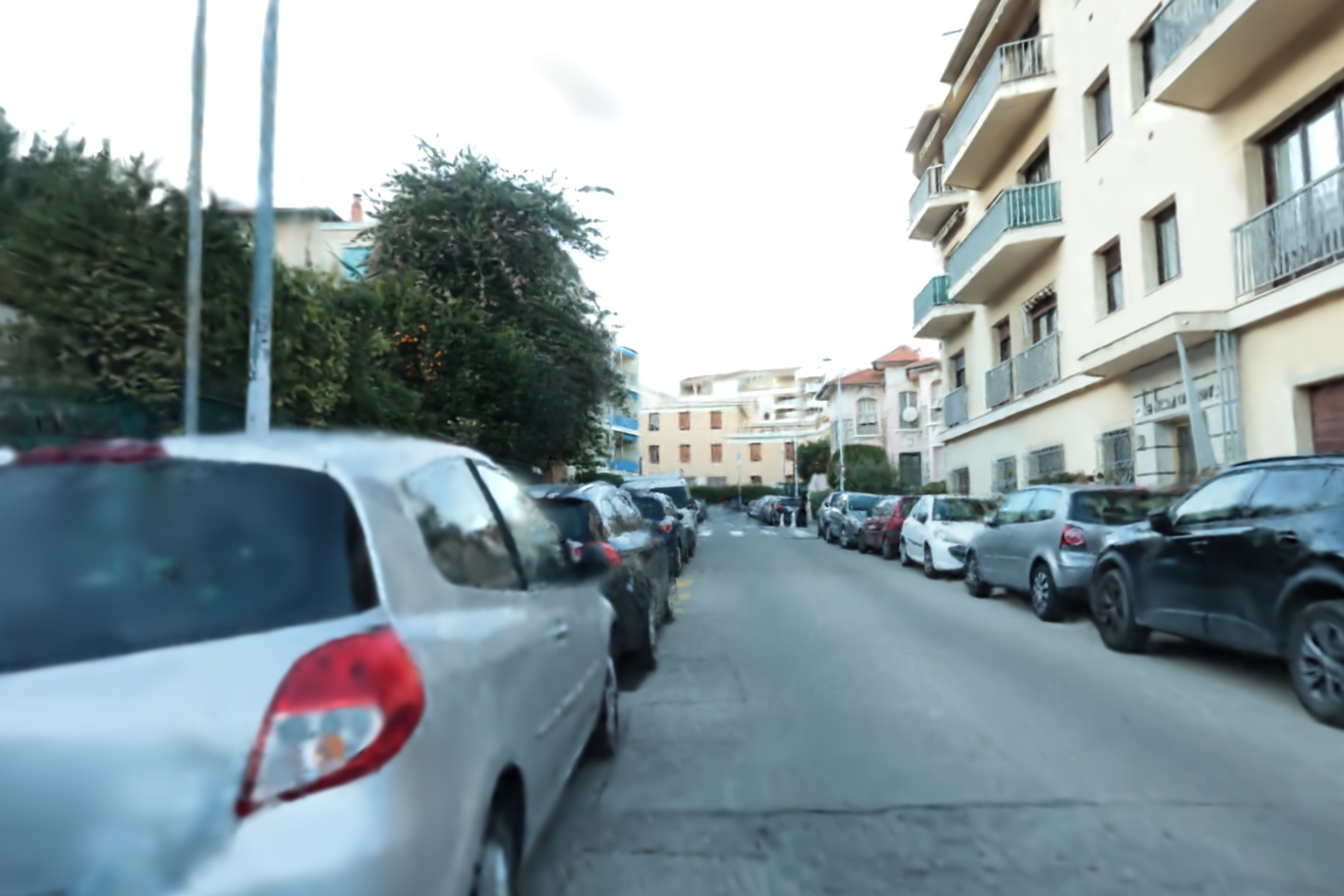}{.475\linewidth}{0:25:03, 21.71 dB} 
\end{tabular}
\vspace{-7pt}
\caption{
\label{fig:compare-large}
	Qualitative comparison of large-scale methods for the three datasets used. 
	\NEW{We show the scene reconstruction time and PSNR.}
}
\end{figure}

\subsection{Pose Estimation Quality}
We evaluate pose estimation quality using the APE and RPE metrics in Tab.~\ref{tab:pose-eval}. 
Our method performs well for the \textsc{MipNeRF360} dataset but has difficulty with \textsc{TUM}.
This is due to the low quality of this video capture, where many frames are blurry, and there is significant rolling shutter which we do not explicitly handle, resulting in bad outlier poses.
SLAM methods are often tuned to perform well on this dataset.
We also compare to Spann3r, a transformer-based approach, whose poses are of lower quality than our method due to substantial pose drift.

\begin{table}
\small
\setlength{\tabcolsep}{1.5pt}
\caption{\label{tab:pose-eval} Pose estimation results for different methods using \NEW{absolute and relative error metrics. The {\colorbox{firstcolor}{best}} and {\colorbox{secondcolor}{second best}} are color coded.}}
\vspace{-7pt}
\begin{tabular}{l|rrrr|rrrr}
\toprule
 & \multicolumn{4}{c|}{\textsc{TUM}} & \multicolumn{4}{c}{\textsc{MipNeRF360}}   \\
 & 
 \footnotesize T.APE$^\downarrow$ & 
 \footnotesize R.APE$^\downarrow$ & 
 \footnotesize T.RPE$^\downarrow$ & 
 \footnotesize R.RPE$^\downarrow$  & 
 \footnotesize T.APE$^\downarrow$ & 
 \footnotesize R.APE$^\downarrow$ & 
 \footnotesize T.RPE$^\downarrow$ & 
 \footnotesize R.RPE$^\downarrow$  \\
\midrule
DROID-Splat &
\first{1.0} & \first{0.033} & \first{1.2} & \first{0.016} & \second{11.7} & \second{0.052} & \second{19.1} & \second{0.074} \\
Photo-SLAM &
\second{9.0} & \second{0.034} & 8.9 & \second{0.021} & 314.0 & 2.016 & 318.9 & 1.213 \\
MonoGS &
33.5 & 0.197 & 23.3 & 0.063 & 315.5 & 2.373 & 278.3 & 0.983 \\
CF-3DGS &
73.1 & 2.817 & 15.0 & 0.187 & 161.5 & 2.777 & 59.5 & 0.197 \\
Spann3r &
89.4 & 0.507 & 33.4 & 0.210 & 32.9 & 0.130 & 42.2 & 0.150 \\
Ours &
40.2 & 0.313 & \second{5.7} & 0.045 & \first{11.4} & \first{0.035} & \first{16.6} & \first{0.047} \\
\bottomrule
\end{tabular}
\end{table}

\NEW{
\subsection{Runtimes}
Tab.~\ref{tab:runtimes} details the runtime of each step of the algorithm on the Garden dataset. 
Each step is executed for every keyframe. Feature detection and extraction are performed for every input frame to determine whether it should be retained as a keyframe.
We process the input images at 40, 4, and 9 FPS and retain 9\%, 86\%, and 31\% of them as keyframes for \textsc{TUM}, \textsc{MipNeRF360}, and \textsc{StaticHikes}, respectively.

\begin{table}
\caption{\label{tab:runtimes} Per keyframe runtime breakdown. 
The pose initialization (Section~\ref{sec:poseinit}) and primitive sampling and placement (Section~\ref{sec:gaussinit}) are relatively efficient and joint pose and Gaussian optimization (Section~\ref{sec:posegauss}) is the longest step.}
\vspace{-7pt}
\NEW{
\begin{tabular}{l|r}
	\toprule
	Step & Time (ms)  \\
	\midrule
	Feature detection and extraction & 3.4 \\
	Matching + Outlier removal & 8.0 \\
	MiniBA incremental & 4.0 \\
	\midrule
	\textbf{Sum pose initialization} & \textbf{15.4} \\
	\midrule
	Monocular depth estimation & 12.9\\
	Triangulation and depth map alignment & 23.2\\
	Dense feature extraction & 2.3\\
	Probability estimation & 0.6\\
	Guided matching & 11.2\\
	\midrule
	\textbf{Sum GS sampling} & \textbf{50.2}\\
	\midrule
	\textbf{Joint optimization} & \textbf{223.8}\\
	\midrule
	\textbf{Sum per keyframe} & \textbf{289.4}\\
	\bottomrule
\end{tabular}
}
\end{table}	
}

Enabling CUDA graphs has a significant impact on the runtime, bringing the bootstrapping optimization time from 625 to 145 and the incremental mini bundle adjustment time from 95 to 4 milliseconds. 
The performance discrepancy shows that setting a fixed-sized problem for mini bundle adjustment is essential to obtain real-time performance in our implementation.

\subsection{Ablations}

We perform ablation studies to illustrate the importance of each step; results are shown in Tab.~\ref{tab:ablations}. First we replace our direct sampling with \NEW{a uniform 0.5 probability of a pixel spawning a Gaussian} (\textsc{NoSampling}). Second, we keep position sampling, but disable shape sampling (\textsc{NoShape}), using the standard 3DGS kNN-based scale initialization. \NEW{Third, we use the monocular depth directly instead of the guided matching (\textsc{NoGuided}).
Finally, we do not refine the poses (\textsc{NoJoint}), i.e., we do not perform joint optimization and just use the initial pose estimation.}

We perform the ablations on two scenes: garden from \textsc{MipNeRF360} and Forest2 from  \textsc{StaticHikes}.

\begin{figure}[h]
    \setlength{\tabcolsep}{1pt}
    \renewcommand{\arraystretch}{0.6}
	\newcommand{\width}{.325\linewidth}
\centering
\begin{tabular}{ccccc}
\textsc{NoSampling} & \textsc{NoShape} & \textsc{OursFull} \\
	\dualspypic{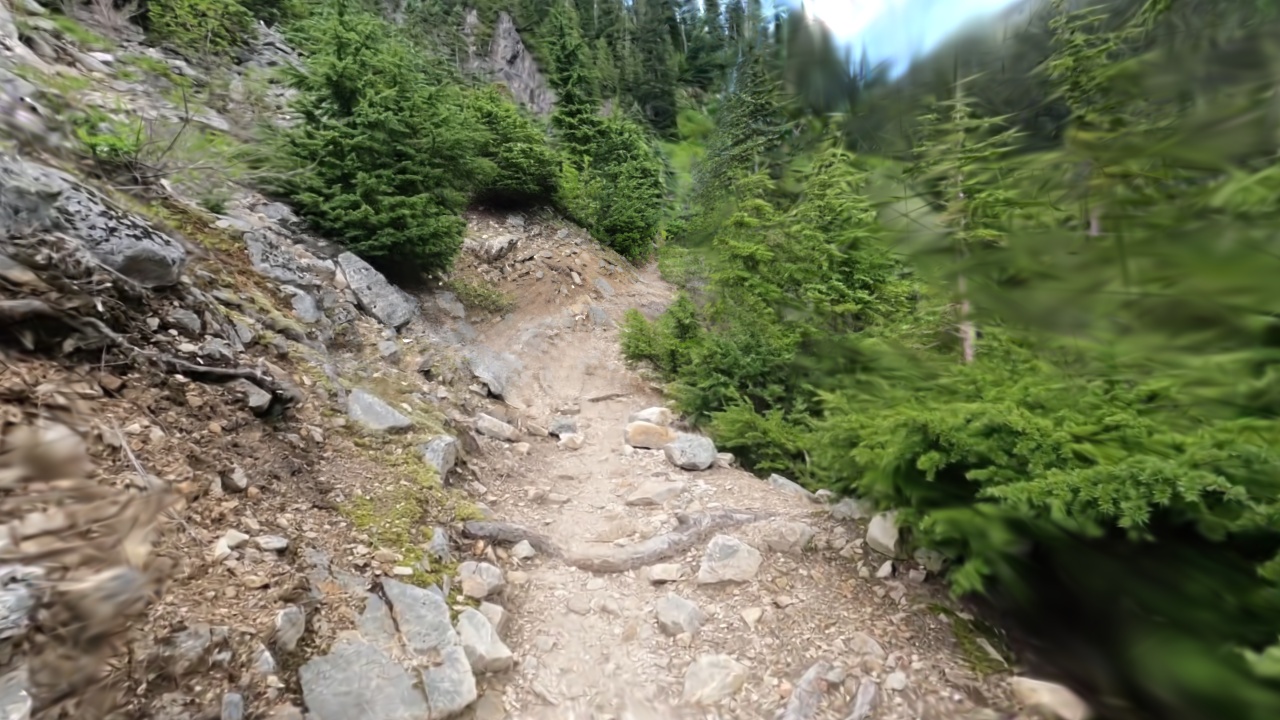}{\width}{0.7*\width,0.35*\width}{0.6pt,-0.253*\width}{0.85*\width,0.15*\width}{\width-0.207pt,-0.253*\width}{3} &
	\dualspypic{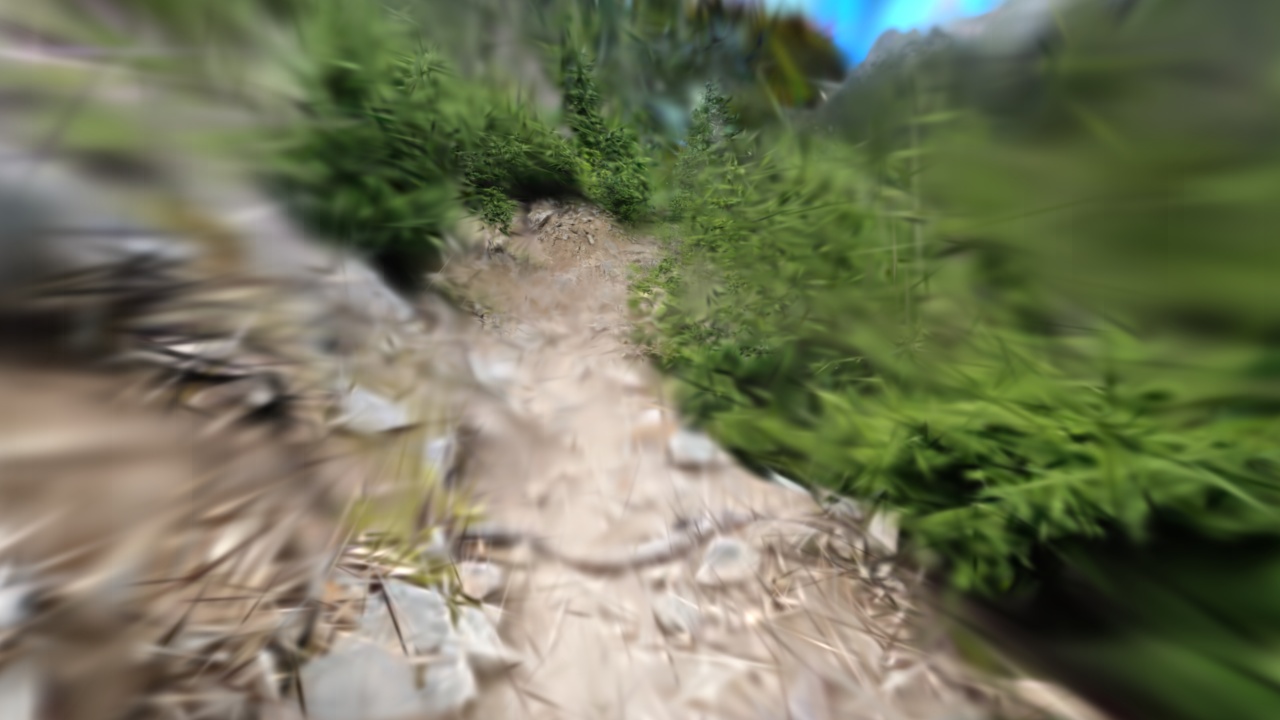}{\width}{0.7*\width,0.35*\width}{0.6pt,-0.253*\width}{0.85*\width,0.15*\width}{\width-0.207pt,-0.253*\width}{3} &
	\dualspypic{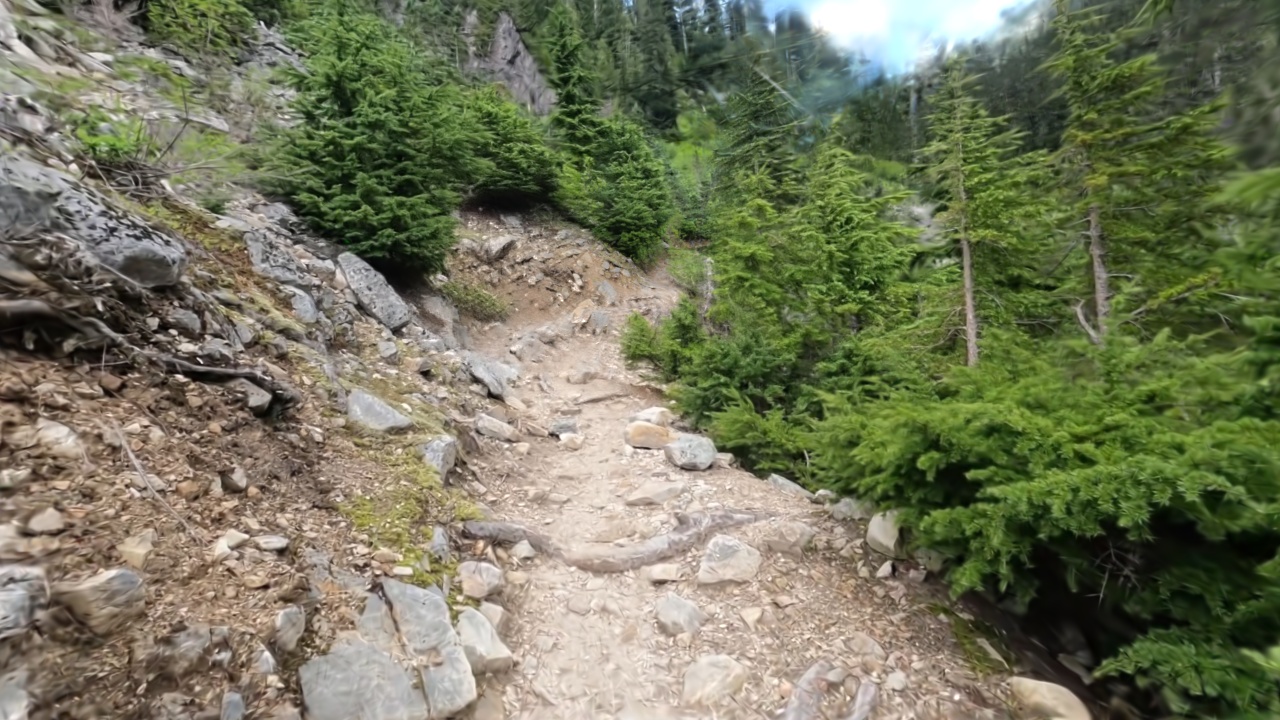}{\width}{0.7*\width,0.35*\width}{0.6pt,-0.253*\width}{0.85*\width,0.15*\width}{\width-0.207pt,-0.253*\width}{3}
\end{tabular}
\vspace{-7pt}
	\caption{\label{fig:ablations}
	Qualitative evaluation of the various components via ablation studies \NEW{on Forest2}.}
\end{figure}

\begin{figure}[h]
    \setlength{\tabcolsep}{1pt}
    \renewcommand{\arraystretch}{0.6}
	\newcommand{\width}{.325\linewidth}
\centering
\begin{tabular}{ccccc}
\textsc{NoGuided} & \textsc{NoJoint} & \textsc{OursFull} \\
	\dualspypic{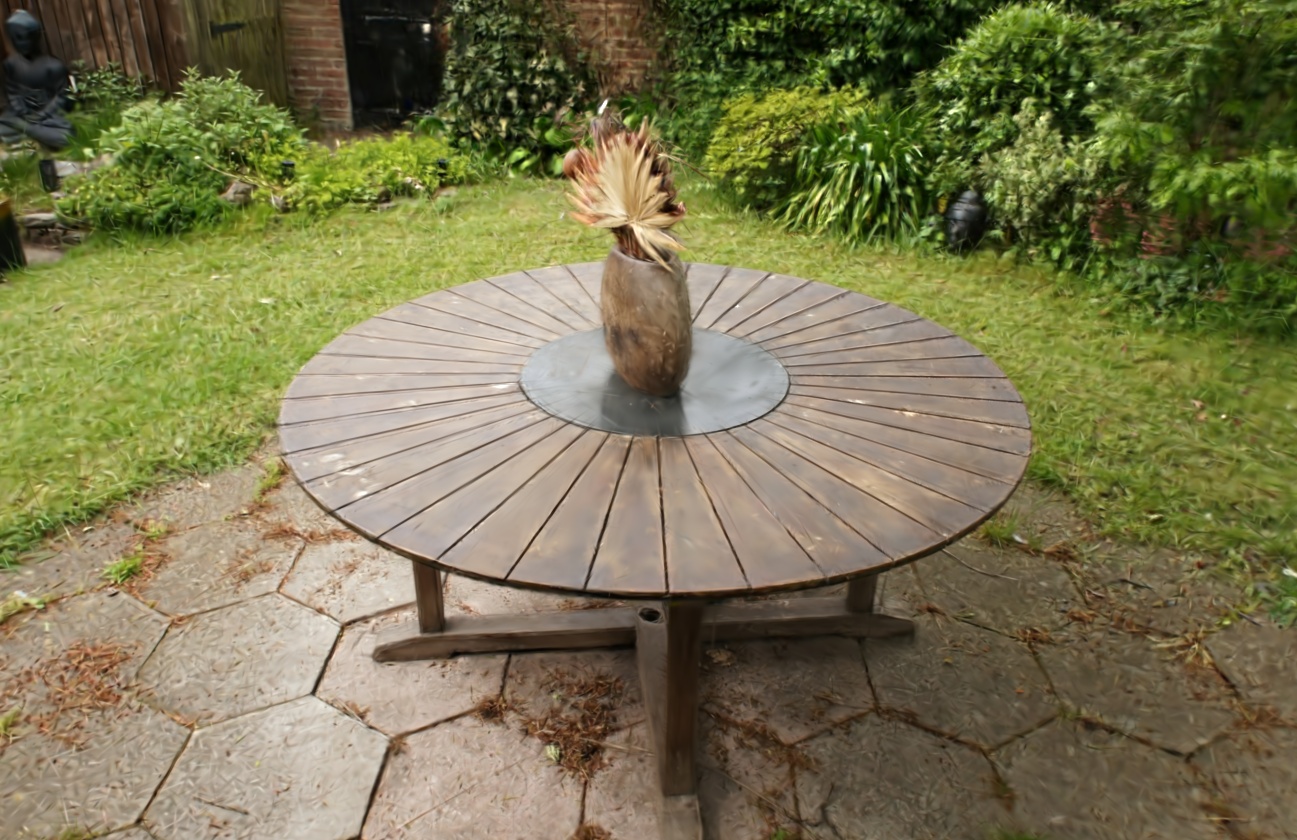}{\width}{0.48*\width,0.505*\width}{0.6pt,-0.253*\width}{0.5*\width,0.16*\width}{\width-0.207pt,-0.253*\width}{4} &
	\dualspypic{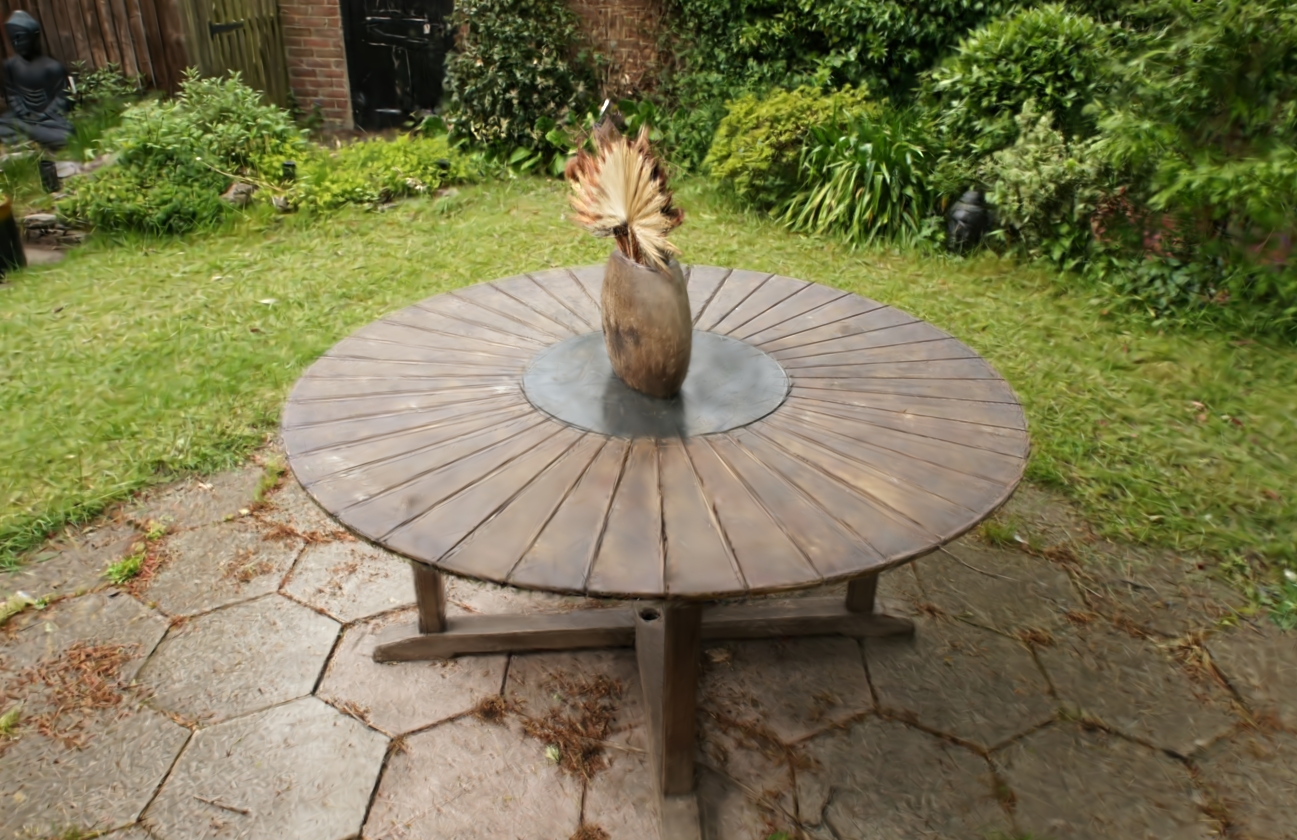}{\width}{0.48*\width,0.505*\width}{0.6pt,-0.253*\width}{0.5*\width,0.16*\width}{\width-0.207pt,-0.253*\width}{4} &
	\dualspypic{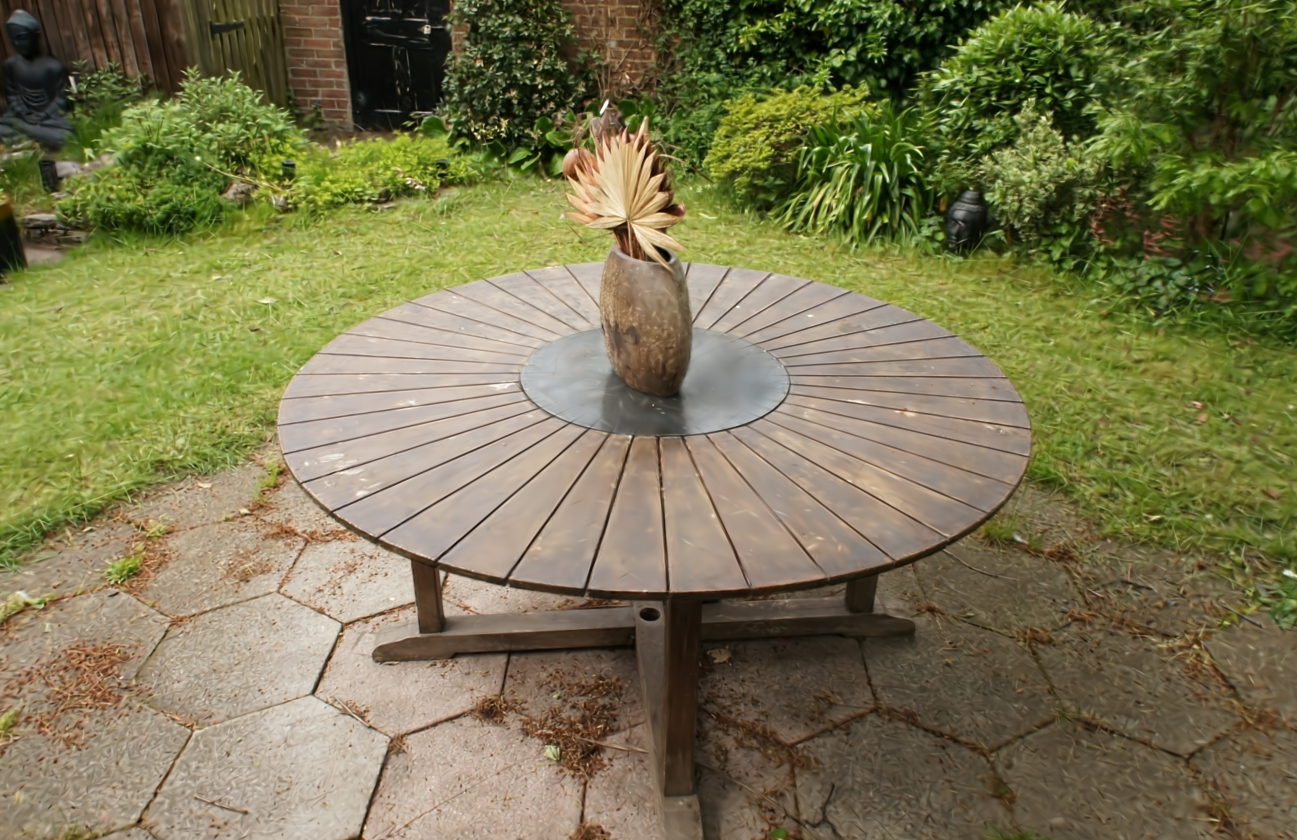}{\width}{0.48*\width,0.505*\width}{0.6pt,-0.253*\width}{0.5*\width,0.16*\width}{\width-0.207pt,-0.253*\width}{4}
\end{tabular}
\vspace{-7pt}
	\caption{\label{fig:ablations2}
	Qualitative evaluation of the various components via ablation studies \NEW{on Garden}.}
\end{figure}

The results show that the each component contributes to the visual quality of our solution. The expectancy-based scale is the most important factor for the \textsc{MipNeRF360} garden scene. Our probability-based sampling
contributes over 1dB in PSNR to overall quality for these cases\NEW{, since it adapts to the input image or the current reconstruction state.} 
In addition, using uniform sampling instead of our Laplacian-based approach is inefficient, bringing the number of primitives from 1.0M to 2.0M on Garden.
\NEW{Figures \ref{fig:ablations} and \ref{fig:ablations2} show qualitative results for the ablations.}

Table~\ref{tab:anchor_ablation} shows the impact of the anchors on a medium-sized scene. 
They have a small positive impact on quality while significantly reducing the peak number of Gaussians rendered. 
For larger scenes, they are required to perform optimization within reasonable GPU memory and runtime\NEW{: they allow the GPU memory usage to stabilize at 22GB after 150 images in CityWalk. This shows that there is (theoretically) no technical limit to the number of images that can be processed}.
Note that, for smaller scenes, the anchor creation criterion is never met, hence not impacting the results. 

\begin{table}
	\caption{\label{tab:anchor_ablation} Impact of the anchors on Forest2. \NEW{Enabling the anchors enhances quality and reduces the maximum number of Gaussians that need to be loaded during optimization.}
	}
	\vspace{-7pt}
	\begin{tabular}{l|cccc}
	\toprule
	 & PSNR$^\uparrow$ & SSIM$^\uparrow$ & LPIPS$^\downarrow$& peak \# GS$^\downarrow$ \\
	\midrule
		Without Anchors & 21.56 & 0.594 & 0.386 & 1432671 \\
		With Anchors & 22.22 & 0.616 & 0.370 & 976688 \\
	\bottomrule
	\end{tabular}
\end{table}

\begin{table}
\caption{\label{tab:ablations} Ablation results.
\textsc{NoSampling} shows results using uniform sampling instead of our direct sampling, 
\textsc{NoShape} without our expectancy-based shape parameters for the primitives,
\textsc{NoGuided} with monocular depth instead of guided matching,
\NEW{\textsc{NoJoint} without pose refinement when optimizing Gaussians}.
}
\vspace{-7pt}
\begin{tabular}{l|ccc}
\toprule
Ablation & PSNR$^\uparrow$ & SSIM$^\uparrow$ & LPIPS$^\downarrow$\\
\midrule
	\textsc{NoSampling} & 21.75 & 0.564 & 0.390 \\
	\textsc{NoShape} & 19.12 & 0.422 & 0.541 \\
	\textsc{NoGuided} & 21.91 & 0.561 & 0.388 \\
	\NEW{\textsc{NoJoint}} & 21.61 & 0.559 & 0.379\\
	\textsc{OursFull} & 23.01 & 0.649 & 0.328 \\
\bottomrule
\end{tabular}
\end{table}

\section{Discussion}
\label{sec:discuss}

Our on-the-fly method for radiance field reconstruction 
provides immediate feedback that is
central in ensuring user-friendly 3D reconstruction. Our method 
will greatly simplify and accelerate 3D capture from photos.
We next discuss limitations and future work.

Our method currently relies on ordered image sequences. 
This is often the case in radiance field captures, but the assumption does not
always hold. For example, several 
scenes in the \textsc{360} dataset~\cite{barron2022mipnerf360} are unordered.
An interesting future avenue of research is the addition of
loop closure that would solve this issue and contribute to an even more
robust solution. Such a solution would allow the user to directly identify
regions that are not well reconstructed, and immediately take more pictures.
This removes one of the most problematic aspects of 3D capture today, where
returning to a capture site to take additional photos is always time-consuming and sometimes logistically impossible. The challenge is how to achieve this and maintain performance.
We also require at least 1000 pixel width resolution for the method to work; given todays cameras, we believe this is not a significant limitation.

To achieve state-of-the-art visual quality, additional optimization is required. While existing optimization schedules can improve quality a bit (Sec.~\ref{sec:nvs-eval}), they are not designed to work with the 3DGS representation we create incrementally with direct sampling. A new approach is required, that takes into account the fact that primitives are already quite dense and well-placed, but is able to escape local minima. This is an exciting direction of future research.

Like many other radiance field methods, we do not explicitly handle casual capture
artifacts such as blur, saturation, lens flare or moving objects or people in
the scene. 
This is visible in the blurry results we obtain for example in the \textsc{TUM} scenes, see Fig.~\ref{fig:compare-slam}, \ref{fig:compare-slam2}.
A plethora of methods have been proposed to address some of these in
3DGS (e.g.,~\cite{sabourgoli2024spotlesssplats,Chen_deblurgs2024}); an interesting avenue of future work is to see how such solutions can be adapted to our framework. The main challenge is incorporating these additional solutions while maintaining performance.

Our method is suitable for a phone/camera application that can take photos and transmit them to a workstation for immediate reconstruction and feedback. Such an application will broaden the impact of our approach.

\section{Conclusion}

We have introduced a novel approach for on-the-fly large-scale pose estimation and 3D reconstruction from casually captured ordered photos, enabling high-quality novel view synthesis with immediate feedback. Our method can handle a wide variety of capture styles, varying from dense SLAM-style video, typical wider radiance field captures all the way to large sequential captures of thousands of images over more than 1km distance.
We have shown that our approach is competitive with the best solutions that specialize in a specific capture style and/or scene
size, and is one of the few method that works for all of them.

Our versatile method is an important step towards real-time 3D capture, with many potential applications in a wide variety of domains. Since our method reduces the computational overhead for pose estimation and optimization from 3DGS allowing immediate feedback, it can only increase the already widespread adoption of radiance fields.

\begin{acks}
This work was funded by the European Research Council (ERC) Advanced Grant NERPHYS, number 101141721 \url{https://project.inria.fr/nerphys/}. The authors are grateful to the OPAL infrastructure of the Universit\'e C\^ote d'Azur for providing resources and support, as well as Adobe and NVIDIA for software and hardware donations. This work was granted access to the HPC resources of IDRIS under the allocation AD011015561 made by GENCI.
B. Kerbl received funding by WWTF (project ICT22-055 - Instant Visualization and Interaction for Large Point Clouds).
Thanks to Peter Hedman for early comments and suggestions, and George Kopanas for proofreading a draft.
\end{acks}

\bibliographystyle{ACM-Reference-Format}
\bibliography{references}

\appendix
\section{Appendix: Implementation Details}

We present various implementation details of the different steps of our method.

\subsection{Initial Pose Estimation}
We describe here more details of the first step in our pipleine.

\paragraph{Feature extraction.}
For faster feature extraction and matching, we run the feature extractor model with half precision and CUDA graphs. 

\paragraph{Bootstrapping.}

We initialize the focal length as 0.7 times the image width, poses as identity and the 3D points with depth $1$. 

We then run $200$ iterations of Levenberg-Marquardt optimization with initial $\lambda=1\cdot10^{-5}$.

We use initial damping $\lambda_{\text{init}} = 10^{-5}$ with factor $\nu=2$ applied at each iteration such that $\lambda_{i+1} = \lambda_i / \nu$ if $\|r_i+1\| < \|r_i\|$, $\lambda_{i+1} = \nu \lambda_i$ otherwise.

\NEW{We employ the Huber loss and discard residuals whose errors exceed the sum of the median and four times the median absolute deviation, ensuring robustness to potential outliers.}

The fixed-size layout also allows us to use CUDA graphs, so all iterations run with a single CPU call, further accelerating computation.

\subsection{Depth for Gaussian Primitives}

We construct a correlation volume with respect to neighboring frames by adapting the dense feature extractor from \cite{ma2022multiview} and applying it to each frame, using half precision and CUDA graphs, similar to the feature detector, to optimize performance. This produces a per-pixel feature map $\mathcal{F}$ for each frame.

Next, we select the most suitable neighboring frame for each pixel position, following the adaptive matching approach in \cite{Meuleman_2021_CVPR}. This adaptive selection ensures that only one neighboring frame per pixel is used, maximizing depth disambiguation while remaining efficient. 

We then construct a candidate depth vector, $z_0, \dots, z_{N-1}$. We restrict the search to a region around the estimated monocular depth: we uniformly sample in inverse depth over the range:
\begin{equation}
    \left[\frac{1}{Z^*} - 10^{-1}, \frac{1}{Z^*} + 10^{-1}\right].
\end{equation}
For each candidate $z_k$, we reproject the pixel $(x, y)$ from the current frame $i$ to the selected neighboring frame $j$, producing coordinates $\left(x^{i\rightarrow j}_k, y^{i\rightarrow j}_k\right)$.

This allows us to build a correlation vector as:
\begin{equation}
    C_k = \left\langle\mathcal{F}_i(x, y), \mathcal{F}_j\left(x^{i\rightarrow j}_k, y^{i\rightarrow j}_k\right)\right\rangle
\end{equation}
where the inner product measures the similarity between feature vectors. We then determine the optimal depth candidate via quadratic fitting.

\subsection{Joint Optimization}

During joint optimization, we initialize all learning rates to a fixed value when the Gaussians are introduced, then when we run \NEW{the sparse Adam optimizer}, we multiply by the decay ratio for each Gaussian processed. 

\subsection{Evaluation Methodology Details}
\NEW{
\paragraph{Using a test/train split}
We aim to optimize poses for all test views without using them to optimize the 3DGS representation. For our method, we enforce keyframe addition for the test images.
Photo-SLAM uses ORB-SLAM3 as its SLAM backend. We modify ORB-SLAM3's keyframing logic to ensure test frames are registered. If the local mapper performing bundle adjustment is busy, the frame is added to the queue. This approach is successful for \textsc{TUM} and \textsc{StaticHikes}, but ORB-SLAM3 registers only about two-thirds of the test images for \textsc{MipNeRF360}. We evaluate the method based on the registered subset.
MonoGS registers all frames in the sequence, providing poses for all test frames regardless of keyframe selection. 
For DROID-Splat, we update the keyframing criteria to always include test frames. In our experiments, DROID-Splat successfully tracked and registered all test frames.
Additionally, all methods were modified to ensure that test frames were not used to optimize the scene; specifically, we skip the Gaussian optimizer for test frames. 

\paragraph{Photo-SLAM runtime}
The time taken by Photo-SLAM's optimization is determined by the frames' timestamps. For \textsc{TUM}, we use the provided timings. For \textsc{StaticHikes} and \textsc{MipNeRF360}, we assume intervals of 0.1s and 0.5s between images, respectively, to maintain a similar runtime order of magnitude as ours.

\paragraph{Focal length}
We use COLMAP's focal length for Photo-SLAM, MonoGS, and COLMAP Free 3DGS, as these methods cannot estimate intrinsics. 
}

\end{document}